\documentclass[10pt,twocolumn,letterpaper]{article}
\usepackage{cvpr}
\usepackage{times}
\usepackage{epsfig}
\usepackage{graphicx}
\usepackage{amsmath}
\usepackage{amssymb}
\usepackage{subfigure}
\usepackage{enumerate}

\usepackage{algorithm}
\usepackage{algorithmicx}
\usepackage{algpseudocode}
\usepackage{color}
\usepackage[breaklinks=true,bookmarks=false]{hyperref}

\cvprfinalcopy % *** Uncomment this line for the final submission

\def\cvprPaperID{2987} % *** Enter the CVPR Paper ID here

\begin{document}

%%% START TRICKS TO SAVE SPACE
\renewcommand{\dblfloatpagefraction}{1}
\renewcommand{\textfraction}{0}
\renewcommand{\dblfloatsep}{5pt}
\renewcommand{\dbltextfloatsep}{1ex}
\renewcommand{\dbltopfraction}{1}
\renewcommand{\intextsep}{5pt}
\renewcommand{\floatpagefraction}{1}
\renewcommand{\floatsep}{5pt}
\renewcommand{\textfloatsep}{1ex}
\renewcommand{\topfraction}{1}
\renewcommand{\abovecaptionskip}{2pt}
\let\OldCaption=\caption
\renewcommand{\caption}[1]{\small\OldCaption{\em#1}}

\newcommand{\AH}[1]{\textcolor{red}{#1}}

%I copied stuff out of art10.sty and modified them to conform to IEEE format
\makeatletter
%as Latex considers descenders in its calculation of interline spacing,
%to get 12 point spacing for normalsize text, must set it to 10 points
\def\@normalsize{\@setsize\normalsize{10pt}\xpt\@xpt
\abovedisplayskip 10pt plus2pt minus5pt\belowdisplayskip
\abovedisplayskip \abovedisplayshortskip \z@
plus3pt\belowdisplayshortskip 6pt plus3pt
minus3pt\let\@listi\@listI}
%need an 11 pt font size for subsection and abstract headings
\def\subsize{\@setsize\subsize{12pt}\xipt\@xipt}
%make section titles bold and 12 point, 2 blank lines before, 1 after
\def\section{\@startsection {section}{1}{\z@}{1.0ex plus
1ex minus .2ex}{.2ex plus .2ex}{\large\bf}}
%make subsection titles bold and 11 point, 1 blank line before, 1 after
\def\subsection{\@startsection {subsection}{2}{\z@}{.2ex
plus 1ex} {.2ex plus .2ex}{\subsize\bf}} \makeatother

% Add the period after section numbers.  Adjust spacing.
\iffalse
\newcommand{\Section}[1]{\vspace{-8pt}\section{\hskip -1em.~~#1}\vspace{-3pt}}
\newcommand{\SubSection}[1]{\vspace{-3pt}\subsection{\hskip -1em.~~#1}
       \vspace{-3pt}}
       \fi
\newcommand{\Section}[1]{\section{\hskip -1em.~~#1}}
\newcommand{\SubSection}[1]{\subsection{\hskip -1em.~~#1}}
\def\@listI{%
 \leftmargin\leftmargini
 \partopsep 0pt
 \parsep 0pt
 \topsep 0pt
 \itemsep pt
 \relax
} \long\def\@makecaption#1#2{
 \vskip -5pt
 \setbox\@tempboxa\hbox{\small{#1\,:\,#2}}
  \ifdim \wd\@tempboxa >\hsize \unhbox\@tempboxa\par \else
  \hbox to\hsize{\hfil\box\@tempboxa\hfil}
\fi \vskip -0.2cm}

% Space for math display
\jot=0pt \abovedisplayskip=3pt \belowdisplayskip=3pt
\abovedisplayshortskip=0pt \belowdisplayshortskip=0pt
% ------
\newcommand\blfootnote[1]{%
	\begingroup
	\renewcommand\thefootnote{}\footnote{#1}%
	\addtocounter{footnote}{-1}%
	\endgroup
}
% ------

%%%%%%%%% TITLE
\title{Accurate Weakly Supervised Deep Lesion Segmentation on CT Scans: Self-Paced 3D Mask Generation from RECIST}

% \author{Jinzheng Cai\\
% University of Florida\\
% Gainesville, FL 32611, USA\\
% {\tt\small jimmycai@ufl.edu}
% \and
% \author{Jinzheng Cai$^{1,2}$\thanks{jimmcai@ufl.edu}~\footnotemark[2]\quad Youbao Tang$^{1}$\thanks{First two author contributed Equally. This work is done during Jinzheng Cai's intern at NIH. Le Lu' work is mostly done when he was an employee of NIH.}\quad Le Lu$^{1,3}$\quad Adam P. Harrison$^{1}$\quad Ke Yan$^{1}$\quad \\
% 	Jing Xiao$^{4}$\quad Lin Yang$^{2}$\quad Ronald M. Summers$^{1}$\\
%	National Institutes of Health, Bethesda, MD, USA$^1$\\
%	University of Florida, Gainesville, FL, USA$^2$\\
%	NVIDIA Corporation, Santa Clara, CA, USA$^3$\\
%	Ping An Insurance (Group) Company of China, Ltd., Shenzhen, PRC$^4$
%}
\author{Jinzheng Cai$^{1,2}$\footnotemark[1]~\footnotemark[2] \quad Youbao Tang$^{1}$\thanks{Indicates equal contribution.} \quad Le Lu$^{1}$\thanks{This work is done during Jinzheng Cai's internship at National Institutes of Health. Le Lu is now with Nvidia Corp (lel@nvidia.com).} \quad Adam P. Harrison$^{1}$ \quad Ke Yan$^{1}$ \quad \\ \vspace{1mm}
Jing Xiao$^{3}$ \quad Lin Yang$^{2}$ \quad Ronald M. Summers$^{1}$\\
National Institutes of Health, Bethesda, MD, 20892, USA$^{1}$ \\
University of Florida, Gainesville, FL, 32611, USA$^{2}$ \\ \vspace{1mm}
Ping An Insurance (Group) Company of China, Ltd., Shenzhen, 510852, PRC$^{3}$ \\ 
jimmycai@ufl.edu \quad \{youbao.tang, le.lu, adam.harrison, ke.yan\}@nih.gov \\
xiaojing661@pingan.com.cn \quad lin.yang@bme.ufl.edu \quad rms@nih.gov
}
\maketitle
%\thispagestyle{empty}

%%%%%%%%% ABSTRACT
\begin{abstract}
Volumetric lesion segmentation via medical imaging is a powerful means to precisely assess multiple time-point lesion/tumor changes. Because manual 3D segmentation is prohibitively time consuming and requires radiological experience, current practices rely on an imprecise surrogate called response evaluation criteria in solid tumors (RECIST). Despite their coarseness, RECIST marks are commonly found in current hospital picture and archiving systems (PACS), meaning they can provide a potentially powerful, yet extraordinarily challenging, source of weak supervision for full 3D segmentation. Toward this end, we introduce a convolutional neural network based weakly supervised self-paced segmentation (WSSS) method to 1) generate the initial lesion segmentation on the axial RECIST-slice; 2) learn the data distribution on RECIST-slices; 3) adapt to segment the whole volume slice by slice to finally obtain a volumetric segmentation. In addition, we explore how super-resolution images ($2\sim5$ times beyond the physical CT imaging), generated from a proposed stacked generative adversarial network, can aid the WSSS performance. We employ the DeepLesion dataset, a comprehensive CT-image lesion dataset of $32,735$ PACS-bookmarked findings, which include lesions, tumors, and lymph nodes of varying sizes, categories, body regions and surrounding contexts. These are drawn from $10,594$ studies of $4,459$ patients. We also validate on a lymph-node dataset, where 3D ground truth masks are available for all images. For the DeepLesion dataset, we report mean Dice coefficients of 93\% on RECIST-slices and $76\%$ in 3D lesion volumes. We further validate using a subjective user study, where an experienced radiologist accepted our WSSS-generated lesion segmentation results with a high probability of 92.4\%. 
\end{abstract}

%%%%%%%%% BODY TEXT
\section{Introduction}

\begin{figure}[t!]
	\begin{center}
		\includegraphics[width=\linewidth]{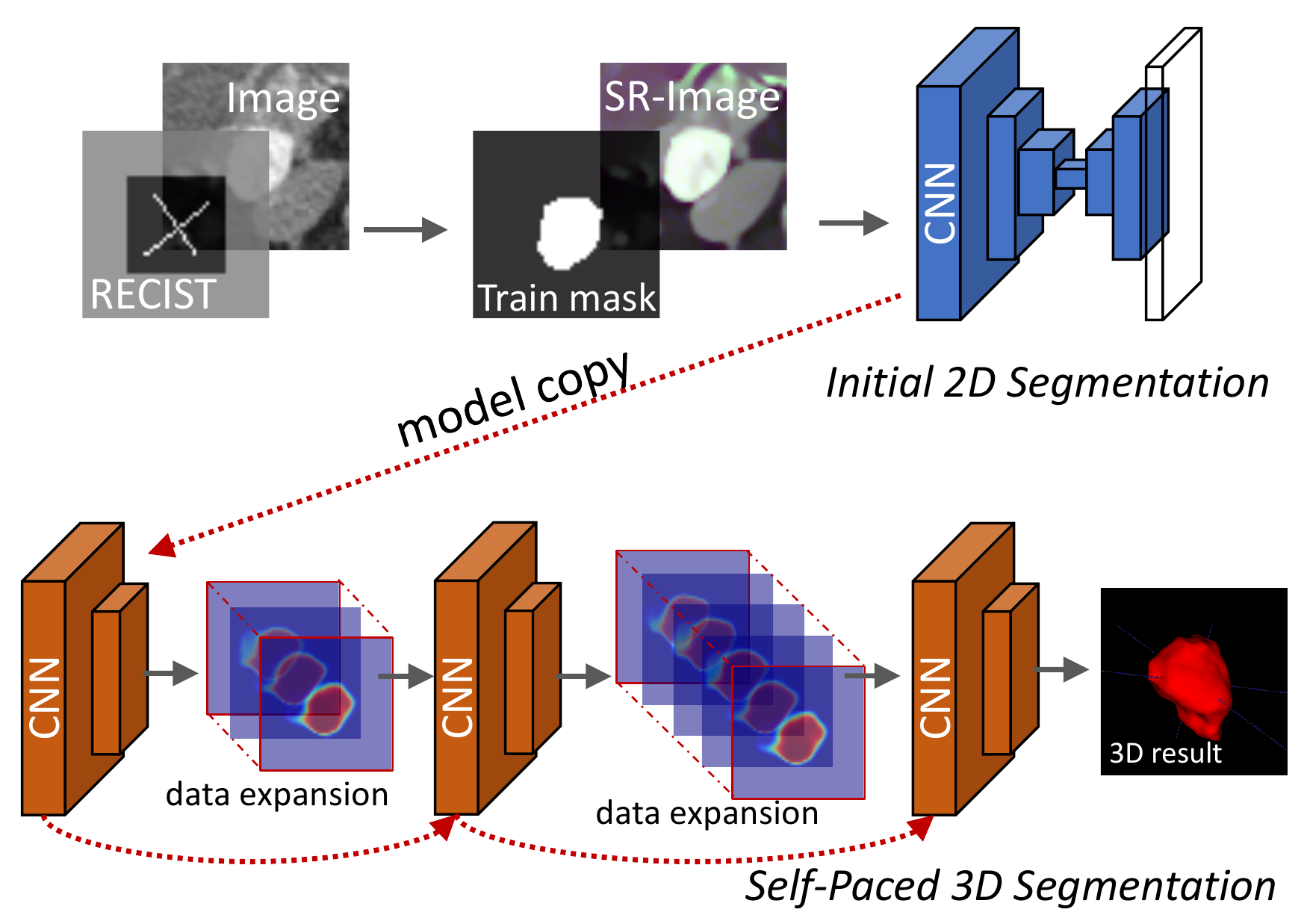}
	\end{center}
	\vspace{-4mm}
	\caption{Overview of the proposed weakly supervised self-paced segmentation with CNN (Sec.\ref{sec:method}) for 3D lesion segmentation.} 
	\label{fig:Self-Paced}
\end{figure}

\begin{figure}[t]
\begin{center}
\setlength{\fboxsep}{0pt}
%% %% \fbox{\includegraphics[width=\linewidth]{figures/overview_small.pdf}} \\[2mm]
 \fbox{\includegraphics[width=0.23\linewidth, height=0.23\linewidth]{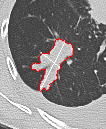}} \hfill
 \fbox{\includegraphics[width=0.23\linewidth, height=0.23\linewidth]{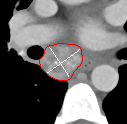}} \hfill
 \fbox{\includegraphics[width=0.23\linewidth, height=0.23\linewidth]{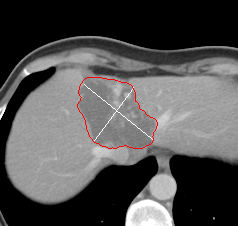}} \hfill
 \fbox{\includegraphics[width=0.23\linewidth, height=0.23\linewidth]{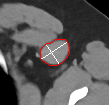}} \\[2mm]
 \fbox{\includegraphics[width=0.23\linewidth, height=0.23\linewidth]{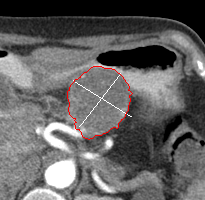}} \hfill
 \fbox{\includegraphics[width=0.23\linewidth, height=0.23\linewidth]{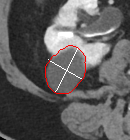}} \hfill
 \fbox{\includegraphics[width=0.23\linewidth, height=0.23\linewidth]{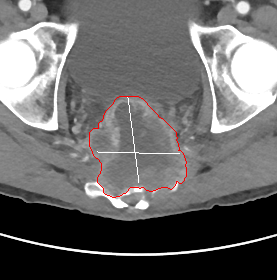}} \hfill
 \fbox{\includegraphics[width=0.23\linewidth, height=0.23\linewidth]{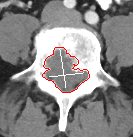}} \hfill
\end{center}
\vspace{-4mm}
\caption{The DeepLesion dataset~\cite{yan_2017_deep-lesion} consists of $8$ categories of lesions. From left to right and top to bottom, the lesion categories lung, mediastinum, liver, soft-tissue, abdomen, kidney, pelvis and bone, respectively. For all images, the RECIST-slices are shown with manually delineated boundary in red and bookmarked RECIST diameters in white.}
\label{fig:data-category}
\end{figure}

Assessing lesion or tumor growth rates of multiple time-point scans of the same patient represents one of the most critical problems in imaging-based precision medicine. To track lesion progression using current clinical protocols, radiologists manually conduct this task using response evaluation criteria in solid tumors (RECIST)~\cite{eisenhauer_2009_recist}  in computed tomography (CT) or magnetic resonance imaging (MRI) scans. A radiologist first selects an axial image slice where the lesion has the longest spatial extent, then he or she measures the diameters of the in-plane longest axis and the orthogonal short axis. Fig.~\ref{fig:data-category} provides some visual examples of RECIST marks provided in DeepLesion dataset \cite{yan_2017_deep-lesion}. Typically, the selected RECIST-slice represents the lesion at its maximum cross area. RECIST is often subjective and prone to inconsistency among different observers, especially when selecting the corresponding slices at different time-points where diameters are measured. However, consistency is critical in assessing actual lesion growth rates, which directly impact patient treatment options. On the other hand, if the volumetric lesion measurements between baseline and follow-ups can be accurately computed and compared, this would avoid the subjective selection of RECIST-slices and allow more holistic and accurate quantitative assessment of lesion growth rates. Unfortunately, full volumetric lesion measurements are too labor intensive to obtain. For this reason, RECIST is treated as the default, but imperfect, clinical surrogate of measuring lesion progression. 

Since the clinical adoption of the RECIST criteria, many modern hospitals' picture archiving and communication systems (PACS) and radiology information systems (RIS) have stored tremendous amounts of lesion diameter measurements linked to lesion CT and MRI images. In this paper, we tackle the challenging problem of leveraging existing RECIST diameters to produce fully volumetric lesion segmentation in 3D. Our approach is a weakly supervised semantic segmentation setup, using convolutional neural networks (CNNs) as the core building block. The overall method flow-chart is given in Fig.~\ref{fig:Self-Paced}. DeepLesion database \cite{yan_2017_deep-lesion} that we exploit is composed of a very large number of $32,735$ significant clinical radiology findings (lesions, tumors, lymph nodes, etc) from $10,594$ studies of $4,459$ patients, bookmarked and measured via RECIST diameters by physicians as part of their day-to-day work. The lesion instances are also clustered into eight different anatomical categories as shown in Fig.~\ref{fig:data-category}. Without loss of generality, here we only consider CT lesion images. 

From any input CT image with RECIST-measured diameters, we aim to segment the lesion region on the RECIST-selected image first in a weakly supervised manner, followed by generalizing the process into other successive slices to finally obtain the lesion's full volume segmentation. More specifically, with the bookmarked long and short RECIST diameters, we initialize the segmentation using unsupervised learning methods (\eg{}, GrabCut \cite{rother_2004_grabcut}). Afterwards we employ an iterative segmentation refinement via a supervised deep neural network model \cite{khoreva_cvpr17_simpledoesit}, which can segment the lesion with good accuracy on the RECIST-slice. Importantly, the resulting CNN lesion segmentation model, trained from all training instances of CT RECIST-slices, can capture the lesion image appearance distribution. Thus, the model is  capable of detecting lesion regions from images off the corresponding RECIST-slice. With more slices segmented, more image data can be extracted and used to fine-tune the deep CNN lesion segmentation model. As such, the proposed weakly supervised segmentation model is a self-paced label-map propagation process, from the RECIST-slice to the whole lesion volume. Therefore, we leverage a large amount of retrospective (yet clinically annotated) imaging data to automatically achieve the final 3D lesion volume measurement and segmentation. %, and no extra human annotation efforts are further needed.

To further enhance our WSSS model, we also investigate the role of image enhancement. A large portion of clinically significant findings in DeepLesion are spatially small ($5\sim10$ mm), which often require significant zooming by radiologists to make precise diameter measurements. To emulate this radiological practice, we develop a stacked generative adversarial network (SGAN)~\cite{ledig2016photo} model to perform resolution magnification, noise reduction, contrast adjustment and boundary enhancement from the original resolution CT images. We stack two GANs together, where the first reduces the noise from original CT image and the second generates a higher resolution image with enhanced boundaries and high contrast. Because the CT modality can only be imaged at a limited range spatial resolution ($\sim 1$ mm/pixel) and super-resolution CT images (\eg{}, $\sim0.2$ mm/pixel) do not physically exist, we train our SGAN on a large quantity of paired low- and high-quality natural images and use transfer learning to apply the model to CT images.

We evaluate the proposed WSSS segmentation method using all lesion categories~\cite{yan_2017_deep-lesion}. To the best of our knowledge, this is the first work to develop a class-agnostic lesion segmentation approach. For quantitative evaluation, we manually annotated 1,000 RECIST-slices and 200 lesion volumes, where the proposed WSSS achieves mean Dice coefficients of $93\%$ and $76\%$, respectively. To compare WSSS against a fully-supervised approach, we also validate on a lymph node (LN) dataset \cite{roth_2014_lymphnode,Seff20142D}, consisting of $984$ LNs with full pixel-wise annotations. Finally, we also conduct a subjective user study, and demonstrate that radiologist accept the WSSS generated lesion masks with high probability of 92.4\%. %Additionally, we employ the lymph node (LN) dataset \cite{roth_2014_lymphnode,Seff20142D}, consisting of 389 mediastinal and 595 abdominal LNs from 176 CT scans with full pixel-wise annotations \cite{n_isabella_miccai2016_lymphnode}. As one subtype lesion, the task of enlarged lymph node segmentation itself can be very challenge even under the fully supervised setting \cite{n_isabella_miccai2016_lymphnode}. LN may share similar appearances with abdomen lesions and soft-tissue lesions. The pixel-wise annotated Lymph-Node dataset \cite{n_isabella_miccai2016_lymphnode} makes it possible to evaluate our weakly supervised method by comparing it with the fully supervised learning performance.
In summary, we present a weakly supervised semantic segmentation approach for accurate lesion segmentation and volume measurements in the wild, using a comprehensive dataset \cite{yan_2017_deep-lesion}. With no extra human annotation efforts required, we convert tens of thousands of recorded RECIST 2D diameter based measurements in PACS/RIS into accurate 3D lesion volume segmentation assessments. %For the rest of this paper, we discuss related work in Sec.\ref{sec:related-work} and describe our proposed self-paced weakly supervised deep segmentation method in Sec.\ref{sec:method}. The segmentation enhancement by conducting cascaded GAN based image super-resolution is also addressed. The extensive experimental evaluations are given in Sec.\ref{sec:exp}, and finally we conclude our work in Sec.\ref{sec:conclusion}.

%!TEX root=../main.tex

%-------------------------------------------------------------------------
\section{Related Work} \label{sec:related-work}
Weakly supervised semantic segmentation is a challenging computer vision task that has drawn considerable interest. Recently promising results have been reported~\cite{d_pap_iccv17_extreneclick,khoreva_cvpr17_simpledoesit,d_lin_cvpr16_scribblesup,g_papandreou_iccv15_wssl,j_dai_iccv15_boxsup,bearman_2015_pointsegment,wei_2015_stc,hong_2017_wsssweb}. Most weakly supervised semantic segmentation methods update pixel labels and CNN models using training iterations. The pixel-level training label is often initialized from weak annotations, \eg{}, bounding boxes of foreground objects. Subsequently, CNNs are trained to capture the initial masks. Although many pixels are possibly initially mislabelled, the CNNs are expected to robustly handle label noise (to some extent) and model the complete data distribution. Once the CNN model converges, its output is used to obtain more accurate pixel labels. Updates between CNN and pixel training labels loop iteratively until no further changes can be measured. However, as observed in \cite{khoreva_cvpr17_simpledoesit}, very high-quality initial pixel labels can allow the CNN model to converge in as few as $1\sim2$ iterations.

Thus, it is essential to have the object mask ``correctly'' initialized and updated. A popular mask initialization uses multiscale combinatorial grouping (MCG)~\cite{j_pt_tpami17_mcg} to group superpixels into plausible object parts, using a bounding box to delineate foreground, background, and uncertain regions during CNN training~\cite{j_dai_iccv15_boxsup, khoreva_cvpr17_simpledoesit}. RECIST diameters are similar to scribbles over foreground objects, and these can be used to coarsely delineate superpixels into foreground and background~\cite{d_lin_cvpr16_scribblesup}. However, empirically MCG does not perform well under the CT imaging modality (compared to natural images) and thus is not suitable for our lesion segmentation task. Semantic edge based methods, like GrabCut$^+$ \cite{khoreva_cvpr17_simpledoesit, d_pap_iccv17_extreneclick}, are also not applicable because unsupervised edge detection methods, such as gPb \cite{j_pt_tpami17_mcg}, suffer from abundant false positives or perform unstably across the whole CT image dataset. %Similarly, GrabCut \cite{khoreva_cvpr17_simpledoesit} and dense Conditional Random Field (DCRF) \cite{g_papandreou_iccv15_wssl} can replace MCG sometimes, or in the extreme case that simply setting the center of the bounding box as foreground (i.e., low recall) where training with high precision object labels is found to be beneficial \cite{khoreva_cvpr17_simpledoesit}, for train CNNs under weak supervision.
%On the contrary, direct pixel-level GrabCut \cite{rother_2004_grabcut, khoreva_cvpr17_simpledoesit} is found to be able to generate the initial object training label maps of high quality, especially when the {\textit{Trimap}} \cite{rother_2004_grabcut} is carefully generated. %But good training label can not be generated for the lesion volume at the very beginning, iterative updating between training label and CNN is still implemented in our work. %DCRF \cite{krah_2012_fully-crf} is successful in natural image segmentation, but in CT images, pixel value in lesion is sometimes similar to objects in the background so that DCRF fails to deliver promising lesion segmentation with many false positive be produced.
Another stream of related work is extending weakly supervised semantic segmentation from static images to video~\cite{hong_2017_wsssweb}. In videos, the same object often retains similar appearance across different temporal frames even though the locations may vary. Using this prior, the training images of object-wise labels can be extracted and propagated across video frames to facilitate CNN training. %However, in our lesion segmentation task, the ImageNet pre-trained appearance CNN models \cite{Deng2009ImageNet,simonyan_2014_vggnet,he_2016_resnet} to detect common object in image sequences are not applicable. 
However lesions exhibit not only positional movements but also drastic morphological changes across CT slices, adding further challenges to lesion volume segmentation from RECIST diameters. Hence, our work differs from prior art by employing a principled \textit{trimap} mask initialization and GrabCut~\cite{rother_2004_grabcut}, coupled with self-paced learning~\cite{kumar_2010_sp-svm} to progressively refine the 3D lesion segmentations. % than processing WSSS in the videos of nature objects.

Recently, deep CNN based image super-resolution methods \cite{dong2016image,kim2016accurate,kim2016deeply,ledig2016photo,sajjadi2016enhancenet,shi2016real,johnson2016perceptual,lai2017deep,taiimage2017,tong2017image,dahl2017pixel} have achieved state-of-the-art performance, due to CNN's capability to learn powerful features and model long-range contextual information from a collection of low resolution images. Mappings between low and high resolution images are learned by minimizing the mean squared error (MSE) loss or perceptual loss \cite{johnson2016perceptual,ledig2016photo,sajjadi2016enhancenet}. Adversarial learning strategies \cite{ledig2016photo} are also employed to train CNN models for better reconstruction of fine details and edges. In this work, we adopt the perceptual loss and adversarial learning to simulate high-resolution images ($\sim0.2$ mm per-pixel) via CNNs directly from the original CT images ($\sim1.0$ mm per-pixel). Unlike other works, we use the enhanced images to improve the WSSS performance.

%No low- and high-resolution image pairs are available for training.

%-------------------------------------------------------------------------
\section{Method} \label{sec:method}

\begin{figure*}[t!]
	\begin{center}
		\includegraphics[width=0.92\linewidth]{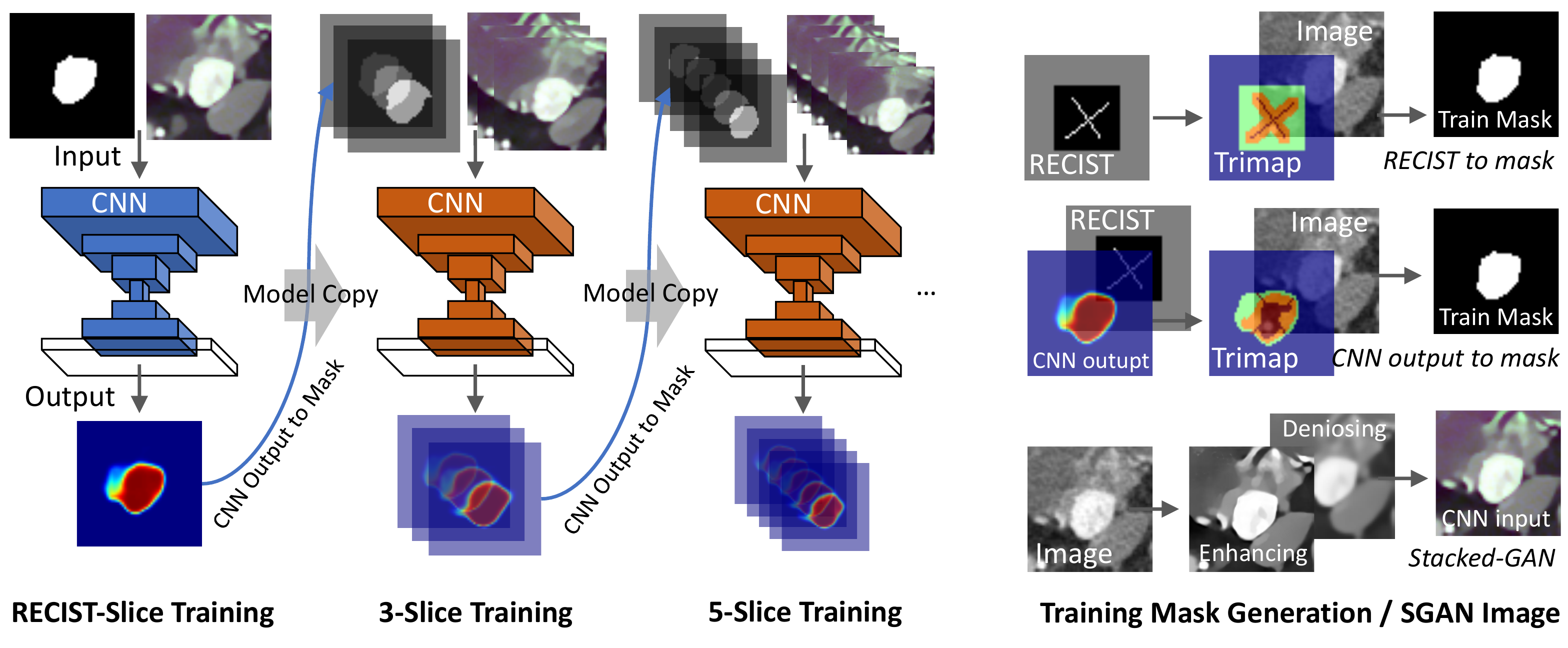}
	\end{center}
	\vspace{-4mm}
	\caption{Our weakly supervised deep lesion segmentation framework. We use CNN output to gradually generate extra training data for self-paced learning. Regions colored with red, orange, green, and blue inside the \textit{trimap} present FG, PFG, PBG, and BG in respective.}
	\label{fig:overview}
\end{figure*}

%-------------------------------------------------------------------------
In DeepLesion dataset \cite{yan_2017_deep-lesion}, manually selected axial CT slices contain RECIST diameters that represent the longest axis and its perpendicular counterpart. Given RECIST diameters as weak supervision, we leverage weakly supervised principles to learn a deep lesion CNN segmentation model using both 2D slice and 3D volumes with no extra pixel-wise manual annotations. Our pipeline for WSSS contains four main steps as follows. %% %%: 1) converting RECIST diameters to training masks, 2) learning a 2D CNN lesion appearance model, 3) expanding the 2D model into 3D segmentation, and optionally 4) using SGAN to generate enhanced images as input. 

\subsection{From RECIST to 2D Training Masks} \label{sec:init-mask}
We denote elements in the DeepLesion dataset \cite{yan_2017_deep-lesion} as {\small $\mathcal{D}=\{(V^i,R^i)\}$} for {\small $i \in \{1,\ldots,N\}$}, where {\small $N$} is the number of lesions, {\small $V^i$} is the CT region of interest, and {\small $R^i$} is the corresponding RECIST diameters. To create the 2D training data for the WSSS model, the image and label map pairs, {\small $X^i$} and {\small $Y^i$}, respectively, must be generated. Together they comprise the set {\small $\mathcal{F}=\{(X^i, Y^i)\}$}, for {\small $i \in \{1,\ldots,N\}$}. For notational clarity, we drop the superscript $i$. {\small $X=V_{r}$} is simply the RECIST-slice, \ie{}, the axial slice that contains {\small $R$}, and $r$ is its slice number. Both GrabCut \cite{rother_2004_grabcut} and densely-connected conditional random fields (DCRF) \cite{krah_2012_fully-crf} can be adopted to produce the training mask $Y$. When GrabCut starts with a good quality initial \textit{trimap} (explained below), we observe that it produces better results, which is consistent with the findings in \cite{khoreva_cvpr17_simpledoesit}.

More specifically, GrabCut is initialized with the image foreground and background regions and produces a segmentation using iterative energy minimization. Good initializations located near the global energy minimum may aid GrabCut's convergence. We use the spatial prior information, provided by {\small $R$}, to compute a high quality initial \textit{trimap} {\small $T$} from {\small $R$}, \ie{}, regions of probable background (PBG), probable foreground (PFG), background (BG), and foreground (FG). Note that unlike the original \textit{trimap} definition~\cite{rother_2004_grabcut}, we define four region types. More specifically, if the lesion bounding box tightly around the RECIST axes is $[w,h]$, a $[2w,2h]$ ROI is cropped from the RECIST-slice. The \textit{trimap} assigns the outer $50\%$ as BG and allocates $10\%$ of the image region, dilated from {\small $R$} as FG. The remaining $40\%$ is divided in half between PFG and PBG based on the distances to FG and BG. Finally, the long/short axes of $R$ roughly clamp the size of GrabCut segmentation, ensuring that the segmentation will not shrink into a small region when the CT image intensity distribution is homogeneous and no clear lesion boundary is presented. Fig.~\ref{fig:overview} visually depicts the mask generation process.

%-------------------------------------------------------------------------
\subsection{CNN Lesion Appearance Model}
Many image classification or segmentation CNN models have been presented in recent works \cite{simonyan_2014_vggnet,he_2016_resnet,huang_2017_densenet,xie_2015_hnn,ronneberger_2015_unet}. Without loss of generality, we use the holistically-nested network (HNN) \cite{xie_2015_hnn} and UNet \cite{ronneberger_2015_unet} as our baselines, both of which provide state-of-the-art, yet straightforward semantic architectures. We represent our CNN models as a mapping function {\small $\hat{Y} = f(X;\theta)$}, with the goal of optimizing {\small $\theta$} to minimize the differences between the current model outputs {\small $\hat{Y}$} and ground truth training labels {\small $Y$}. Although the GrabCut masks are imperfect, we use them as {\small $Y$} for the next step of CNN model training, expecting that the CNN will generalize well even with considerable label noise. Thus, we aim to minimize
\begin{equation} \label{eq:convnet}
L(\mathcal{F}; \theta) = \frac{1}{N} \sum\limits_{i=1}^{N} \frac{\sum\limits_{m\in Y^i} H (\hat{y}^{(i,m)},y^{(i,m)})}{|Y^i|} \textrm{,}
\end{equation}
where {\small $m$} is the pixel index inside {\small $Y$}, {\small $|Y|$} is the cardinality of {\small $Y$}, and {\small $H(\cdot)$} is the cross-entropy function. 

%-------------------------------------------------------------------------
\subsection{Self-Pacing for Volume Segmentation} \label{sec:3d-wsss}
For obtaining 3D volume segmentations, we follow a similar strategy as with the RECIST slices, except in the 3D case, we must infer $R$ for off-RECIST slices and also incorporate inferences from the CNN model. These two priors are used together for self-paced CNN training.

\textbf{3D RECIST Estimation:} A simple way to generate off-RECIST-slice  {\small $R$} diameters is to take advantage of the fact that RECIST-slice {\small $R$} lies on the maximal cross-sectional area of the lesion. The rate of reduction of off-RECIST-slice endpoints is then calculated by their relative distance to the intersection point of the major and minor axes. Estimated 3D RECIST endpoints are then projected from the actual RECIST endpoints by Pythagorean theorem using physical Euclidean distance.

\textbf{Trimap from CNN Output:} Different from Sec.~\ref{sec:init-mask}, \textit{trimap} generation now takes both the CNN output {\small $\hat{Y}$} and the estimated RECIST {\small $\hat{R}$} as inputs: {\small $\hat{T} = trimap(\hat{Y},\hat{R},R)$}. The {\small $\hat{Y}$} is first binarized by adjusting the threshold so that it covers at least $50\%$ of {\small $R$}'s pixels. Regions in {\small $\hat{Y}$} that associate with high foreground probability values and overlap with {\small $R$} will be set as FG together with {\small $R$}. Similarly, regions with high background probabilities and that have no overlap with {\small $R$} will be assigned as BG. The remaining pixels are left as uncertain using the same distance criteria as in the 2D mask generation case and fed into GrabCut for lesion segmentation. In the limited cases where the CNN fails to detect any foreground regions, we generate {\small $\hat{T}$} solely from the estimated RECIST {\small $\hat{R}$}. We observe that when possible, GrabCut initialized by the CNN output performs better than being solely initialized from {\small $\hat{R}$}.

\textbf{Self-Paced CNN Training:} To generate 3D lesion segmentations from 2D RECIST annotations, we train the CNN model in a self-paced or self-taught mechanism. It begins by learning the lesion appearances via CNN on RECIST-slices. After the model converges, {\small $Y=GrabCut(trimap(\hat{Y},\hat{R},R),X)$} is used to generate the training masks on successive slices so that it can be further fine-tuned on more newly ``harvested'' images. We progressively expand the extent of the harvested images, to include more CT slices along the longitudinal axis. Taking one lesion volume for example, the CNN is first trained on its RECIST slice {\small $X$} until convergence, we then apply this CNN model to slices {\small $[V_{r-1},V_{r+1}]$} to compute the predicted probability maps {\small $[\hat{Y}_{r-1},\hat{Y}_{r+1}]$}. Given these probability maps, we create \textit{trimaps} {\small $[\hat{T}_{r-1},\hat{T}_{r+1}]$} to employ GrabCut refinement \cite{khoreva_cvpr17_simpledoesit} on these slices. The GrabCut method consequently generates the training labels that are used by the CNN model in the next training round on the slices {\small $[V_{r-2},V_{r+2}]$} slices. As this procedure proceeds iteratively, we can gradually obtain the converged lesion segmentation result in 3D. We summarize our WSSS method in Algorithm~\ref{alg:wsss} and visually depict its process in Fig.~\ref{fig:overview}. Note that it is possible that {\small $GrabCut(\cdot)$} gives no or very few foreground areas or pixels if the initial \textit{trimap} and CT image are of bad quality. In these cases, we set up foreground, background and ignored uncertain pixels by combining the {\small $\hat{R}$} and the CNN output. CNN predicted foreground regions that overlap with {\small $\hat{R}$} are assigned as foreground; predicted background regions that have no overlap with RECIST are set as background; with the remaining pixels ignored during training.

\begin{algorithm}[t!] 
	\caption{Weakly Supervised Self-Paced CNN Training} \label{alg:wsss}
	\textbf{Input:} $\mathcal{D}={(V^i,R^i)}$ for $i \in \{1, \ldots N\}$  \hfill $\triangleright$ \textit{Deep-Lesion dataset}
	\textbf{Input:} \textit{3D RECIST Estimation:} $\hat{R}^i$ \\
	\textbf{Input:} \textit{Maximum iteration:} $K$, \textit{initial CNN model:} $\theta$.
	\begin{algorithmic}[1]
		\State $X^i = V^i_r$ \hfill $\triangleright$ \textit{RECIST-Slice extraction}
		\State $Y^i = GrabCut(trimap(R^i), X^i)$ \hfill $\triangleright$ \textit{Initial Train-Mask}
		\State $\mathcal{F} = \{(X^i,Y^i)\}$ \hfill $\triangleright$ \textit{RECIST-slices training set}
		\State $\theta^0 := \min L(\mathcal{F},\theta)$ \hfill $\triangleright$ \textit{Training of initial 2D segmentation}
		\For{$k=1$ to $K$}
		\State $\mathcal{F} = \emptyset$
		\For {$\tau=-k$ to $k$}
		\State $\hat{Y}^i_{r+\tau} = f(V^i_{r+\tau}, \theta^{k-1})$, \hfill $\triangleright$ \textit{CNN inference}
		\State $\hat{T}_{r+\tau} = trimap(\hat{Y}^i_{r+\tau},\hat{R}^i_{r+\tau},R^i)$,
		\State $Y^i_{r+\tau} = GrabCut(\hat{T}_{r+\tau}, V^i_{r+\tau})$,
		\State $\mathcal{F} = \mathcal{F} \cup \{(V^i_{r+\tau}, Y^i_{r+\tau})\}$.
		\EndFor
		\State $\theta^k := \min L(\mathcal{F},\theta^{k-1})$, \hfill $\triangleright$ \textit{CNN training with multi-slices}
		\EndFor \\
		\Return $\theta^K$ \hfill $\triangleright$ \textit{Return trained CNN model}
	\end{algorithmic}
\end{algorithm}
%! TEX root=../main.tex
\subsection{Stacked-GAN Assisted Segmentation}

We next investigate improving lesion segmentation accuracy by enhancing the quality and resolution of the input CT images via GANs. A direct application of super-resolution generative adversarial networks (SRGANs) \cite{ledig2016photo} on CT images produces many artificial, noisy edges and even deteriorates the final lesion segmentation outcome. Spurred by this observation, we propose a two-stage stacked GAN (SGAN) process that first reduces image noise and then performs object boundary enhancement. More specifically, each stage is conducted using an independent SRGAN \cite{ledig2016photo}. During model inference, the output of the first GAN will be fed as input to the second GAN. For lesion segmentation, we use both the denoised and enhanced outputs from SGAN and the original CT image as three-channel inputs since they may contain complementary information.

\textbf{Synthesized Training Data:} Normally the SGAN or SRGAN models are trained with pairs of low- and high-resolution images. This can be obtained easily in natural images (by down-sampling). However physical CT images are imaged by medical scanners at roughly fixed in-plane resolutions of $\sim1$ mm per-pixel and CT imaging at ultra-high spatial resolutions does not exist. 
For the sake of SGAN training, we leverage transfer learning using a large-scale synthesized natural image dataset: DIV2K \cite{agustsson_2017_cvpr_workshops} where all images are converted into gray scale and down-sampled to produce training pairs. For the training of the denoising GAN, we randomly crop {\small $32\times32$} sub-images from distinct training images of DIV2K. White Gaussian noise at different intensity variance levels {\small $\sigma_i\in(0,50]$} are added to the cropped images to construct the paired model inputs. For training the image-enhancement GAN, the input images are cropped as {\small $128\times128$} patches and we perform the following steps: 1) down-sample the cropped image with scale {\small $s\in[1,4]$}, 2) implement Gaussian spatial smoothing with {\small $\sigma_s\in(0,3]$}, 3) execute contrast compression with rates of {\small $\kappa\in[1,3]$}, and 4) conduct up-sampling with the scale {\small $s$} to generate images pairs.

To fine-tune using CT images, we process $28,000$ training RECIST slices using the current trained SGAN and select a subset of up to $1,000$ slices by subjectively inspecting the CT super-resolution results. The selected CT images are subsequently added to the training for the next round of SGAN fine-tuning. This iterative process finishes when no more visual improvement could be observed.

%-------------------------------------------------------------------------
\section{Experiments} \label{sec:exp}
%\subsection{Experimental Set Up} \label{sec:exp-sets}
\textbf{Datasets:} The DeepLesion dataset \cite{yan_2017_deep-lesion} is composed of $32,735$ bookmarked CT lesion instances (with RECIST measurements) from $10,594$ studies of $4,459$ patients. Lesions have been categorized into the $8$ subtypes of lung, mediastinum, liver, soft-tissue, abdomen, kidney, pelvis, and bone. To facilitate quantitative evaluation, we segmented $1,000$ testing lesion RECIST-slices manually. Out of these $1000$, $200$ lesions ($\sim3,500$ annotated slices) are fully segmented in 3D as well. Additionally, we also employ the lymph node (LN) dataset\footnote{\url{https://wiki.cancerimagingarchive.net/display/Public/CT+Lymph+Nodes}} \cite{roth_2014_lymphnode,Seff20142D}, which consists of $389$ mediastinal and $595$ abdominal LNs from $176$ CT scans with complete pixel-wise annotations. Enlarged LNs are a lesion subtype and producing accurate segmentations is quite challenging even with fully supervised learning \cite{n_isabella_miccai2016_lymphnode}. Importantly, the fully annotated LN dataset can be used to evaluate our WSSS method against an upper performance limit, by comparing results with a fully supervised approach \cite{n_isabella_miccai2016_lymphnode}. %LN may share similar appearances with abdomen lesions and soft-tissue lesions. %We also study a smaller dataset, the Lymph-Node \cite{roth_2014_lymphnode} \footnote{\url{https://wiki.cancerimagingarchive.net/display/Public/CT+Lymph+Nodes}}, which have been fully annotated in pixel-wise accuracy for enlarged lymph-node in 176 CT volumes. As a subtype of soft-tissue lesion, lymph nodes shares {\textbf{similar visual appearance}} with Deep-Lesions. Working on lymph node segmentation helps us to understand the proposed WSSS method in comprehensive.

{\bf Pre-processing:} For the LN dataset, annotation masks are converted into RECIST diameters by measuring the major/minor axes after localizing the $5$ successive axial slices nearest to the largest LN cross-section. For robustness, up to $20\%$ random noise is injected in the RECIST diameter lengths to mimic the uncertainty of manual annotation by radiologists. For both datasets, based on the location of RECIST bookmarks, CT ROIs are cropped at two times the extent of the lesion's longest diameters so that sufficient visual context is preserved. The dynamic range of each lesion ROI is then intensity-windowed properly using the CT windowing meta-information in \cite{yan_2017_deep-lesion}. The LN dataset is separated at the patient level, using the split of $80\%$ and $20\%$ for training and testing, respectively. For the DeepLesion dataset \cite{yan_2017_deep-lesion}, we use $28,000$ lesion volumes for training and the rest for testing.

{\bf Evaluation:} The mean dice similarity coefficient (mDICE), {\small $DICE=2TP/(2TP+FP+FN)$} and the pixel-wise precision and recall scores are used to evaluate the quantitative segmentation accuracy. We measure volumetric similarity (VS) {\small $VS=1-(FN-FP)/(2TP+FP+FN)$} and averaged Hausdorff distance (AVD) to better capture and evaluate the small but vital changes near object boundaries \cite{t_abdel_bmc15_metrics}.

%Ellipse fitting can be used to approximate the initial foreground lesion segmentation on the RECIST slice so that (weak) supervision is applicable during model training. As tested on Lymph-Node and annotated images in Deep-Lesion, the measuring score of weak supervision is of strong correlation with score from manual ground truth. With strong mDICE, the weak mDICE is of 0.86 and 0.98 Pearson correlation for Deep-Lesion and Lymph-Node in respective. Weak supervision enables model evaluation when we process new lesion data from wild.

Our baseline CNN model is the holistically-nested network (HNN), originally proposed for natural image edge detection \cite{s_xie_iccv15_hed}, which has been adapted successfully for lymph node \cite{n_isabella_miccai2016_lymphnode}, pancreas \cite{j_cai_miccai17_pancreas,r_holger_arxiv16_pancreas}, and lung segmentation \cite{Harrison_2017}. In all experiments, CNN training is implemented in Tensorflow~\cite{tensorflow2015-whitepaper} and Tensorpack~\footnote{\url{https://github.com/ppwwyyxx/tensorpack}} with Adam Optimizer and initialized from the pre-trained ImageNet model \cite{s_xie_iccv15_hed}. The learning rate is set as {\small $5\times10^{-5}$} and it drops to {\small $1\times10^{-5}$} when the model training-validation plot plateaus.

% !TEX root = ../main.tex

\begin{table}[t!]
\begin{center}
\caption{\textbf{Methods to Generate Training Label from RECIST:} pixel-wise precision, recall, and mean DICE (mDICE) are reported with standard deviation ($\pm$stdv.). 5 different setups are compared, 1) RECIST: dilated RECIST, 2) DCRF: dense CRF, 3) GrabCut: uses only RECIST bbox, 4) GrabCut$^i$: uses bbox and interior foreground, 5) GrabCut-R: uses bbox and dilated RECIST. See Sec.~\ref{sec:init-mask} and experiment settings for details.} \vspace{3mm}
\label{tab:mask-init}
{\small
\begin{tabular}{l c c c}
\hline
Method & Recall & Precision & mDICE \\
\hline \hline
\multicolumn{4}{c}{Lymph Node} \\ %%% on Fold 1
\hline
RECIST     & 0.35$\pm$0.09  & \textbf{0.99$\pm$0.05}  & 0.51$\pm$0.09 \\
DCRF      & 0.29$\pm$0.20  & 0.98$\pm$0.05  & 0.41$\pm$0.21 \\
GrabCut    & 0.10$\pm$0.25  & 0.32$\pm$0.37  & 0.11$\pm$0.26 \\
GrabCut$^i$& 0.53$\pm$0.24  & 0.92$\pm$0.10  & 0.63$\pm$0.17 \\
GrabCut-R  & \textbf{0.83$\pm$0.11}  & 0.86$\pm$0.11  & \textbf{0.83$\pm$0.06} \\
\hline
\multicolumn{4}{c}{Deep Lesion RECIST-Slice (Testing Images)} \\
\hline
RECIST     & 0.39$\pm$0.13  & \textbf{0.92$\pm$0.14}  & 0.53$\pm$0.14 \\
DCRF      & 0.72$\pm$0.26  & 0.90$\pm$0.15  & 0.77$\pm$0.20 \\
GrabCut    & 0.62$\pm$0.46  & 0.68$\pm$0.44  & 0.62$\pm$0.46 \\
GrabCut$^i$& \textbf{0.94$\pm$0.11}  & 0.81$\pm$0.16  & 0.86$\pm$0.11 \\
GrabCut-R  & \textbf{0.94$\pm$0.10}  & 0.89$\pm$0.10  & \textbf{0.91$\pm$0.08} \\
\hline
\end{tabular}
}
\end{center}
\end{table}

%% %% \begin{table}[t!]
%% %% \begin{center}
%% %%   {\small
%% %%   \begin{tabular}{|l|c|c|c|}
%% %%   \hline
%% %%   Method & Recall & Precision & F1-score \\
%% %%   \hline \hline
%% %%   \multicolumn{4}{|c|}{LNo-MCA Fold 1} \\
%% %%   \hline
%% %%   RECIST-A & 0.34$\pm$0.10  & 0.94$\pm$0.05  & 0.49$\pm$0.10 \\
%% %%   RECIST-B & 1.00$\pm$0.01  & 0.25$\pm$0.10  & 0.39$\pm$0.11 \\
%% %%   GrabCut  & 0.51$\pm$0.27  & 0.92$\pm$0.07  & 0.61$\pm$0.20 \\
%% %%   GC-Conv  & 0.58$\pm$0.27  & 0.92$\pm$0.07  & 0.67$\pm$0.20 \\
%% %%   \hline \hline
%% %%   \multicolumn{4}{|c|}{DLe-MCA Testing Images} \\
%% %%   \hline
%% %%   RECIST-A & 0.45$\pm$0.14  & 0.91$\pm$0.16  & 0.59$\pm$0.15 \\
%% %%   RECIST-B & 1.00$\pm$0.03  & 0.16$\pm$0.06  & 0.28$\pm$0.08 \\
%% %%   GrabCut  & 0.69$\pm$0.28  & 0.88$\pm$0.16  & 0.74$\pm$0.22 \\
%% %%   GC-Conv  & 0.88$\pm$0.16  & 0.86$\pm$0.13  & 0.85$\pm$0.13 \\
%% %%   \hline
%% %%   \end{tabular}
%% %%   }
%% %% \end{center}
%% %% \caption{Methods to generate training mask from RECIST: RECIST with distance transform; GrabCut use RECIST as trimap; GC+Conv, ConvNet firstly trained on RECIST
%% %% and its outputs is then used to generate trimap for GrabCut.}
%% %% \label{tab:mask-init}
%% %% \end{table}

{\bf Label Map Initializations:} Both GrabCut \cite{khoreva_cvpr17_simpledoesit,rother_2004_grabcut} and densely-connected conditional random fields (DCRF) \cite{krah_2012_fully-crf} are extensively evaluated for initializing training label maps from RECIST diameters. We test two alternatives to the \textit{trimap} approach explained in Sec.~\ref{sec:init-mask}, which we denote GrabCut-R. Both alternatives are based on a tight bounding box (bbox), which matches the extent of the lesion RECIST marks with $25\%$ padding (against the lesion's spatial extent). The first alternative (GrabCut), sets the area outside bbox as background and area inside bbox as probable foreground. The second alternative (GrabCut$^i$) sets the central $20\%$ bbox region as foreground, regions outside the bbox as background, and the rest as uncertain. This is similar to the setting of bbox$^i$ in \cite{khoreva_cvpr17_simpledoesit}. We also test DCRF, using bbox$^i$ as the unary potentials and intensities to compute pairwise potentials \cite{krah_2012_fully-crf}. We empirically found that DCRF is moderately sensitive to parameter variations. The optimal DCRF performance is reported in Table~\ref{tab:mask-init}. Finally, we also report results when we directly use the RECIST diameters, but dilated to $20\%$ of bbox area, which unsurprisingly produces the best precision, but at the cost of very low recall. However, as can be seen in the table, GrabCut-R significantly outperforms all alternatives, demonstrating the validity of our mask initialization process. 

% !TEX root = ../main.tex

\begin{table}[t!]
\begin{center}
\caption{\textbf{Initial Masks to Train the CNN:} all results are reported as (mDICE$\pm$std.). For CNN training, GC-Mask uses GrabCut-R as mask initialization, whereas RECIST uses the \textit{trimap} of Sec.~\ref{sec:init-mask}. The performance achieved by the fully supervised baseline of HNN \cite{xie_2015_hnn} and UNet \cite{ronneberger_2015_unet} is denoted as Full-Sup. CNN-GC is the result post-processed by GrabCut using the CNN outputs to initialize the \textit{trimap}.} \vspace{3mm}
\label{tab:2d-convnet}
{\small
\begin{tabular}{l c c } %c|}
\hline
Method & CNN & CNN-GC \\ \hline \hline
\multicolumn{3}{c}{Lymph-Node} \\ \hline
Full-Sup (HNN) & \textbf{0.710$\pm$0.18} & \textbf{0.845$\pm$0.06}  \\
RECIST         & 0.614$\pm$0.17 & 0.844$\pm$0.06 \\
GC-Mask        & 0.702$\pm$0.17 & 0.844$\pm$0.06  \\
\hline
\multicolumn{3}{c}{Deep-Lesion RECIST-Slice} \\ \hline
Full-Sup (UNet)& 0.728$\pm$0.18 & 0.838$\pm$0.16  \\
Full-Sup (HNN) & 0.837$\pm$0.16 & 0.909$\pm$0.10  \\
RECIST         & 0.644$\pm$0.14 & 0.801$\pm$0.12  \\
GC-Mask        & \textbf{0.906$\pm$0.09} & \textbf{0.915$\pm$0.10}  \\
\hline
\end{tabular}
}
\end{center}
\end{table}

{\bf CNN Training under Different Levels of Supervision:} Following Sec. \ref{sec:init-mask}, there are three ways of generating initial lesion masks on the RECIST-slice: using the \textit{trimap} from the dilated RECIST (ignoring uncertain regions in training), GrabCut-R (processed from the \textit{trimap}) and the full pixel-wise manual annotation (when available). Using the LN dataset \cite{roth_2014_lymphnode,Seff20142D} with manual ground truth, HNNs trained from  these three initializations of label maps achieve 61\%, 70\%, and 71\% mDICE scores, with increased levels of supervision respectively. No extra segmentation refinement options like adaptive sample mining, or weighted training loss are applied. This observation demonstrates the robustness and effectiveness of using GrabCut-R label map initialization, which only performs slightly worse than the fully annotated pixel-wise masks. On the DeepLesion~\cite{yan_2017_deep-lesion} testset of $1,000$ annotated testing RECIST-slices, HNN trained on GrabCut-R initialization outperforms the deep model learned from dilated RECIST maps by a margin of 25\% in mDICE (90.6\% versus 64.4\%). GrabCut post-processing further improves the results from 90.6\% to 91.5\%. We also aim to demonstrate that our WSSS approach, trained on a large quantity of weakly supervised or ``imperfectly-labeled'' object masks, can outperform fully-supervised models trained on less data. To do this, we separated the $1,000$ annotated testing images into five folds and report the mDICE scores over 5-fold cross-validation using fully-supervised HNN~\cite{xie_2015_hnn} and UNet~\cite{ronneberger_2015_unet} models. Impressively, the $90.6\%$ dice score of WSSS considerably outperforms the fully supervised HNN and UNet mDICE scores of $84\%$ and $73\%$, respectively. This demonstrates the importance of training CNNs from a large-scale ``imperfectly-labeled'' dataset. All the results are described in Table~\ref{tab:2d-convnet}. 

%!TEX root=../main.tex

\begin{figure}
\centering
\includegraphics[width=.9\linewidth]{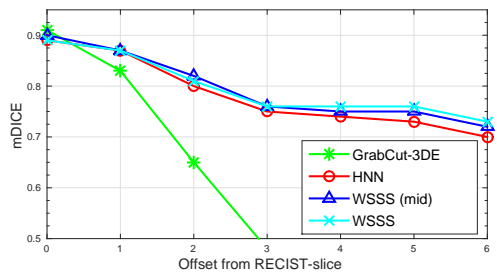}
\caption{\textbf{Volume Segmentation with Offsets:} The x-axis represents the voxel distance of axial slices to the RECIST-Slice. GrabCut-3DE is GrabCut segmentation performed with 3D RECIST estimation. HNN is trained on the RECIST-slice. WSSS (mid) is the HNN self-paced with $3$ axial slices, and WSSS is the HNN further self-paced with $5$ axial slices.}
\label{fig:offset}
\end{figure}

%!TEX root=../main.tex

\begin{figure}
\centering
\mbox{%
\begin{minipage}[b]{0.49\linewidth}
\centering
\scriptsize \hspace{2mm} RECIST-Slice PR-Curve
\includegraphics[width=\linewidth,height=0.85\linewidth]{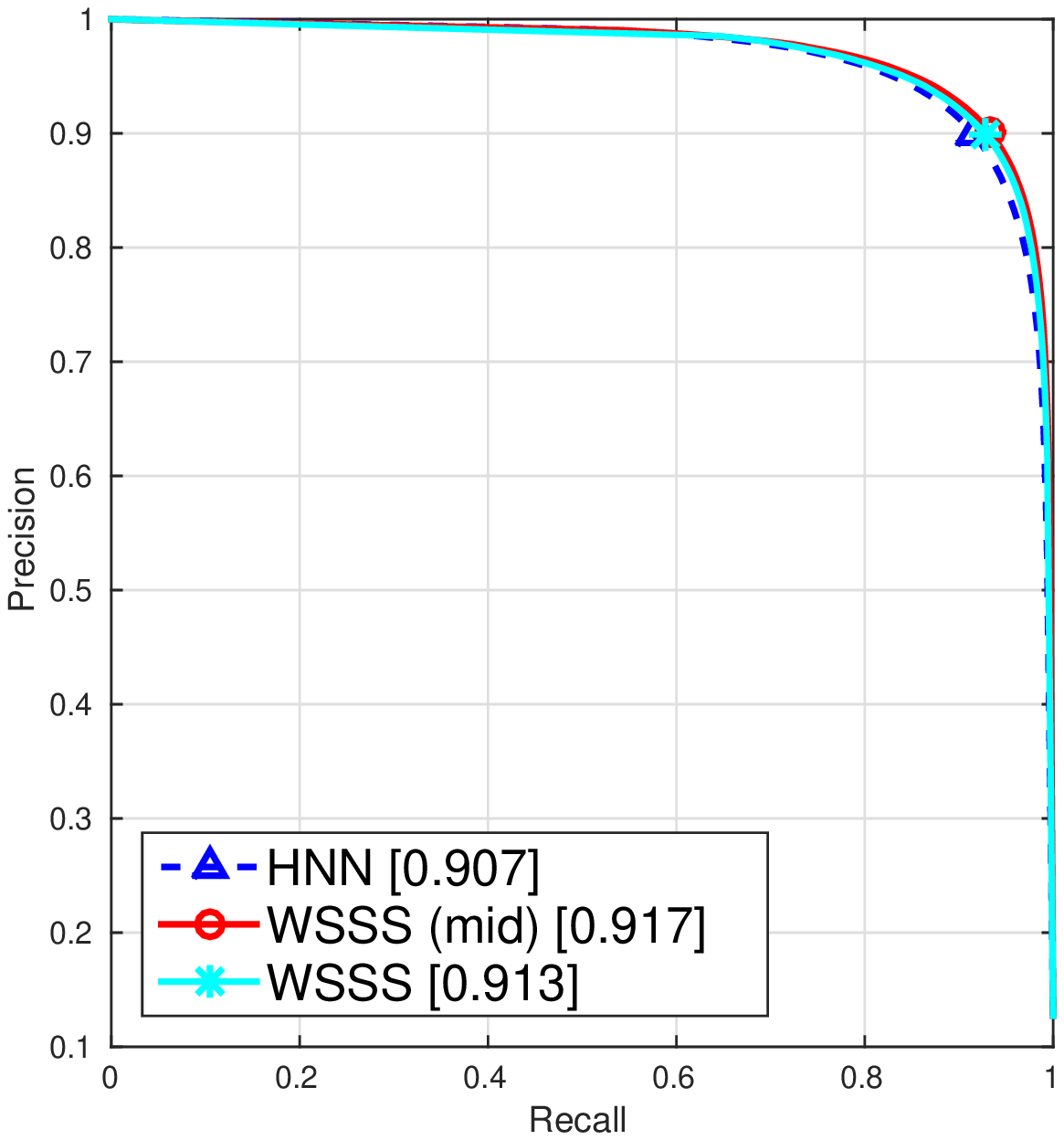}
\end{minipage}
\hspace{2mm}
\begin{minipage}[b]{0.49\linewidth}
\centering
\scriptsize \hspace{2mm} Lesion Volume PR-Curve
\includegraphics[width=\linewidth,height=0.85\linewidth]{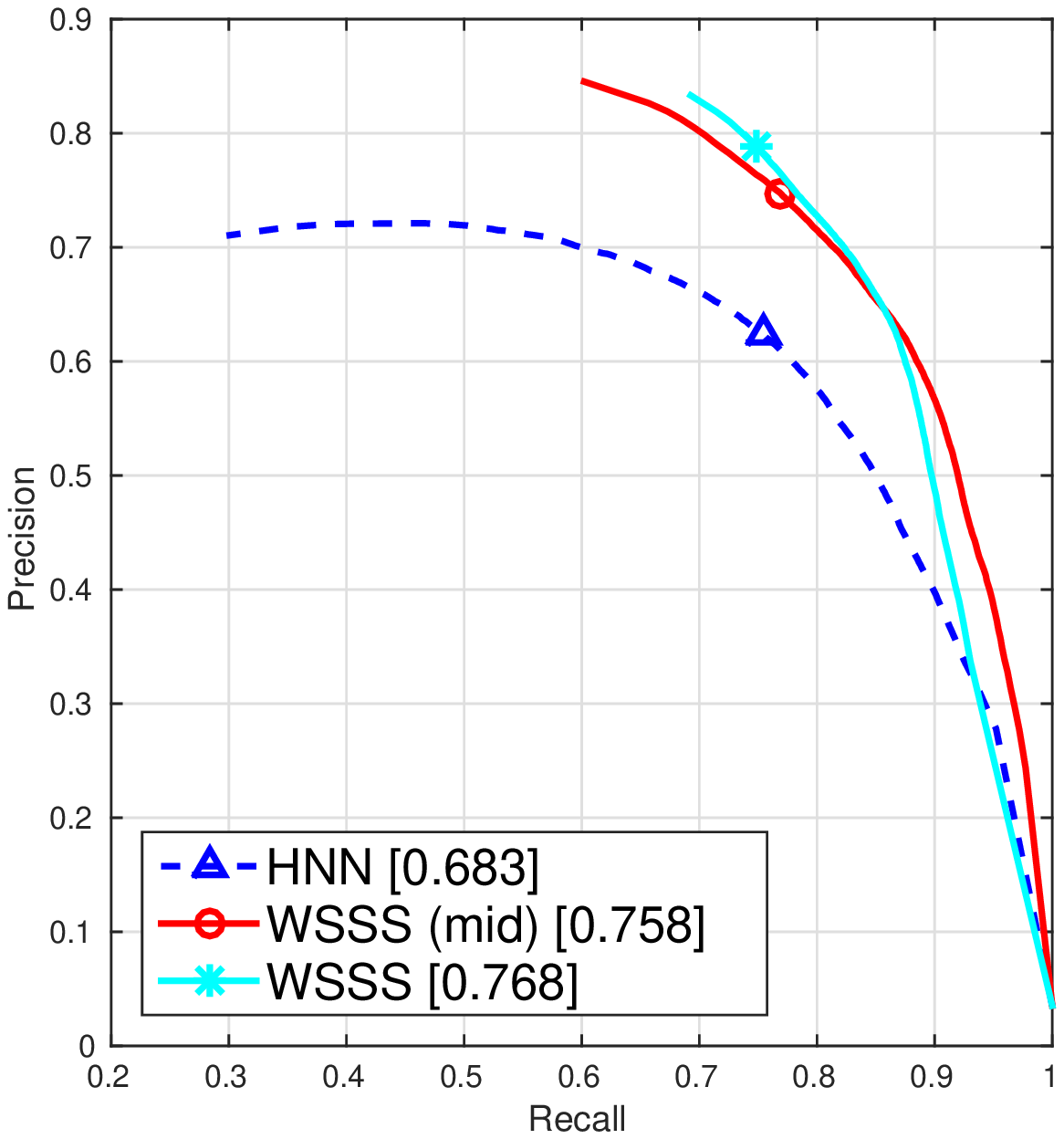}
\end{minipage} \hspace{3mm}}
\caption{\textbf{Dataset Precision-Recall:} The precision and recall curve of CNN output on 1,000 testing RECIST slices, and 200 lesion volumes. CNN, WSSS (mid), and WSSS share the same definition as in Fig.\ref{fig:offset}, and its F-measure is presented in brackets in the legend.}
\label{fig:pr_curves}
\end{figure}

% !TEX root = ../main.tex

\begin{table}[t!]
\begin{center}
\caption{{\bf Ablation Study of Stacked-GAN Assisted Lesion Segmentation} with and w/o Super-Resolution (SR).} 
\label{tab:Ablation-convnet}
{\small
\begin{tabular}{l c c} %c|}
\hline
Metric 		& with SR & w/o SR \\ \hline
Recall 		& 0.911$\pm$0.097 & \textbf{0.933$\pm$0.095} \\
Precision   & \textbf{0.940$\pm$0.091} & 0.893$\pm$0.111 \\
AVD 		& \textbf{0.189$\pm$1.030} & 0.230$\pm$1.070 \\
VS 			& \textbf{0.951$\pm$0.067} & 0.942$\pm$0.073 \\
DICE 		& \textbf{0.920$\pm$0.082} & 0.906$\pm$0.089 \\
\hline
\end{tabular}
}
\end{center}
\end{table}

%% %%  GC-MASK-INIT:  & 0.933$\pm$0.095  & 0.893$\pm$0.111  & 0.906$\pm$0.089  & 0.230$\pm$1.070  & 0.942$\pm$0.073  \\
%% %%  GC-MASK-INIT-SR: & 0.911$\pm$0.097  & 0.940$\pm$0.091  & 0.920$\pm$0.082  & 0.189$\pm$1.030  & 0.951$\pm$0.067  \\
%!TEX root=../main.tex

\begin{table*}[t!]
\begin{center}
\caption{\textbf{Category-Wise RECIST-Slice Segmentation Comparison.} HNN-SR and HNN-GC denotes HNN augmented with super resolution images and using GrabCut post-processing, respectively, whereas HNN-SR-GC uses both enhancements.}
{\scriptsize
\label{tab:lesion-2d}
\begin{tabular}{l c c c c c c c c c}
\hline
Method & Bone & Abdomen & Mediastinum & Liver & Lung & Kidney & Soft-Tissue & Pelvis & Total \\ \hline
%% %% Test 2D     & 26   & 190     & 152         & 82    & 403  & 57     & 92          & 79     & - \\ \hline
GrabCut     & 0.88$\pm$0.07  & 0.92$\pm$0.07  & 0.88$\pm$0.09  & 0.86$\pm$0.13  & 0.92$\pm$0.08  & 0.93$\pm$0.06  & 0.93$\pm$0.07  & 0.91$\pm$0.08  & 0.91$\pm$0.09 \\
HNN     & 0.88$\pm$0.06  & 0.91$\pm$0.09  & 0.89$\pm$0.08  & 0.85$\pm$0.15  & 0.91$\pm$0.09  & 0.93$\pm$0.06  & 0.93$\pm$0.06  & 0.91$\pm$0.07  & 0.91$\pm$0.09 \\
HNN-SR  & 0.89$\pm$0.06  & 0.93$\pm$0.09  & \textbf{0.91$\pm$0.08}  & 0.88$\pm$0.14  & 0.92$\pm$0.07  & \textbf{0.94$\pm$0.05}  & 0.94$\pm$0.05  & \textbf{0.92$\pm$0.08}  & 0.92$\pm$0.08 \\
HNN-GC  & 0.90$\pm$0.08  & 0.92$\pm$0.11  & 0.90$\pm$0.08  & 0.87$\pm$0.14  & 0.92$\pm$0.11  & 0.93$\pm$0.11  & 0.94$\pm$0.06  & 0.92$\pm$0.07  & 0.92$\pm$0.10 \\
HNN-SR-GC  & \textbf{0.92$\pm$0.07}  & \textbf{0.93$\pm$0.06}  & 0.90$\pm$0.08  & \textbf{0.88$\pm$0.11}  & \textbf{0.94$\pm$0.07}  & 0.94$\pm$0.06  & \textbf{0.95$\pm$0.06}  & \textbf{0.92$\pm$0.08}  & \textbf{0.93$\pm$0.07} \\
\hline
\end{tabular}
}
\end{center}
\end{table*}

{\bf 3D Segmentation:} In Fig.~\ref{fig:offset}, we show the segmentation results on 2D CT slices arranged in the order of voxel-distance with respect to the RECIST-selected slice. GrabCut with 3D RECIST estimation (GrabCut-3DE) produces good segmentations ($\sim$91\%) on the RECIST-slice, but can degrade to 55\% mDICE when the off-slice distance raises to 4. This is mainly because 3D RECIST approximation often is not a robust and accurate estimation across slices. %using the unreliable \textit{trimaps} initialized from the 3D RECIST estimation. 
In contrast, the HNN trained using RECIST-slices generalizes well with large slice offsets. It still gains $>70$\% of mDICE even when the offset distance range as high as $6$. However, performance is further improved at higher slice offsets when using self-paced learning with $3$ axial slices, \ie{}, WSSS (mid), and even further when using the full self-paced learning with $5$ axial slices, \ie{}, WSSS.  These results demonstrate the value of using our self-paced learning approach to generalize beyond 2D RECIST slices into full 3D segmentations. Fig.~\ref{fig:pr_curves} further demonstrates the model improvements from self-paced learning using precision-recall curves. Unsurprisingly, the WSSS schemes do not provide much improvement on the 2D RECIST slices; however, significant improvements are garnered when considering the full 3D segmentations, again demonstrating the benefits of WSSS to achieve clinically useful volumetric lesion segmentations. The final 3D segmentation results are tabulated in Table~\ref{tab:lesion-3d}.

%!TEX root=../main.tex

\begin{table*}[t!]
\begin{center}
\caption{\textbf{Category-Wise Lesion 3D Segmentation Comparison.} GrabCut-3D and GrabCut-3DE denotes GrabCut uses the 2D RECIST and the estimated 3D RECIST, respectively, to initialize the trimap. WSSS is the HNN self-paced with 5 axial slices, and WSSS-GC denotes WSSS using GrabCut post-processing. mDICE ($\pm$ stdv.) scores are presented across all $8$ lesion categories.}
\label{tab:lesion-3d}
{\scriptsize
\begin{tabular}{l c c c c c c c c c}
\hline
Method & Bone & Abdomen & Mediastinum & Liver & Lung & Kidney & Soft-Tissue & Pelvis & Total \\ \hline
%% %% Test3D       & 13 & 27 & 16 & 25 & 27 & 22 & 41 & 30 & - \\ \hline
GrabCut-3D   & 0.221$\pm$0.13 & 0.294$\pm$0.23  & 0.268$\pm$0.15  & 0.358$\pm$0.18  & 0.192$\pm$0.18  & 0.340$\pm$0.16  & 0.364$\pm$0.15  & 0.234$\pm$0.16 & 0.292$\pm$0.18 \\
GrabCut-3DE  & 0.654$\pm$0.08 & 0.628$\pm$0.20  & 0.693$\pm$0.15  & 0.697$\pm$0.15  & 0.667$\pm$0.14  & 0.747$\pm$0.15  & 0.726$\pm$0.13  & 0.580$\pm$0.14 & 0.675$\pm$0.16 \\
HNN       	 & 0.666$\pm$0.11 & 0.766$\pm$0.12  & 0.745$\pm$0.11  & 0.768$\pm$0.07  & 0.742$\pm$0.15  & 0.777$\pm$0.07  & \textbf{0.791$\pm$0.08}  & \textbf{0.736$\pm$0.08} & 0.756$\pm$0.11 \\
WSSS         & \textbf{0.685$\pm$0.10}  & 0.766$\pm$0.14  & \textbf{0.776$\pm$0.10}  & \textbf{0.773$\pm$0.06}  & 0.757$\pm$0.15  & \textbf{0.800$\pm$0.06}  & 0.780$\pm$0.10  & 0.728$\pm$0.09 & 0.762$\pm$0.11 \\
WSSS-GC      & 0.683$\pm$0.12 & \textbf{0.774$\pm$0.15}  & 0.771$\pm$0.07  & 0.765$\pm$0.08  & \textbf{0.773$\pm$0.15}  & 0.800$\pm$0.08  & 0.787$\pm$0.10  & 0.722$\pm$0.10 & \textbf{0.764$\pm$0.11} \\ \hline
\end{tabular}
}
\end{center}
\end{table*}

%!TEX root=../main.tex

\begin{figure*}[t]
\begin{center}
\setlength{\fboxsep}{0pt}
 \fbox{\includegraphics[width=0.11\linewidth, height=0.11\linewidth]{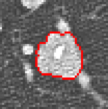}} \hfill
 \fbox{\includegraphics[width=0.11\linewidth, height=0.11\linewidth]{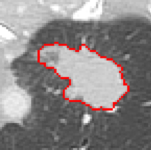}} \hfill
 \fbox{\includegraphics[width=0.11\linewidth, height=0.11\linewidth]{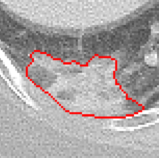}} \hfill
 \fbox{\includegraphics[width=0.11\linewidth, height=0.11\linewidth]{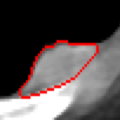}} \hfill
 \fbox{\includegraphics[width=0.11\linewidth, height=0.11\linewidth]{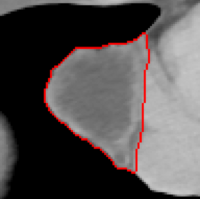}} \hfill 
 \fbox{\includegraphics[width=0.11\linewidth, height=0.11\linewidth]{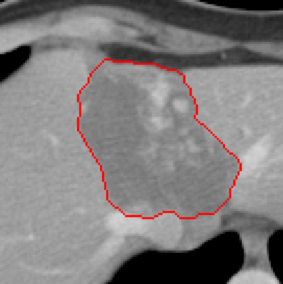}} \hfill
 \fbox{\includegraphics[width=0.11\linewidth, height=0.11\linewidth]{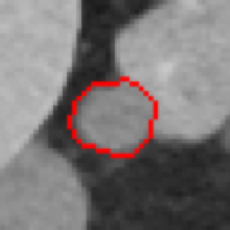}} \hfill
 \fbox{\includegraphics[width=0.11\linewidth, height=0.11\linewidth]{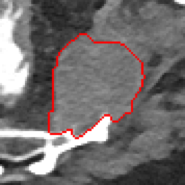}} \\[2mm]
 \fbox{\includegraphics[width=0.11\linewidth, height=0.11\linewidth]{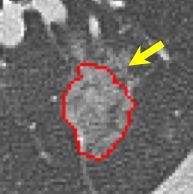}} \hfill
 \fbox{\includegraphics[width=0.11\linewidth, height=0.11\linewidth]{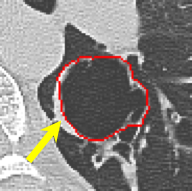}} \hfill
 \fbox{\includegraphics[width=0.11\linewidth, height=0.11\linewidth]{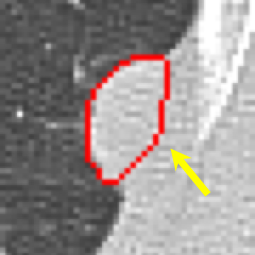}} \hfill
 \fbox{\includegraphics[width=0.11\linewidth, height=0.11\linewidth]{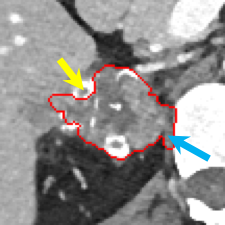}} \hfill
 \fbox{\includegraphics[width=0.11\linewidth, height=0.11\linewidth]{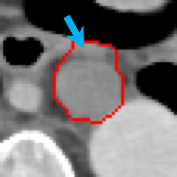}} \hfill
 \fbox{\includegraphics[width=0.11\linewidth, height=0.11\linewidth]{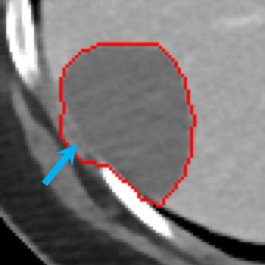}} \hfill
 \fbox{\includegraphics[width=0.11\linewidth, height=0.11\linewidth]{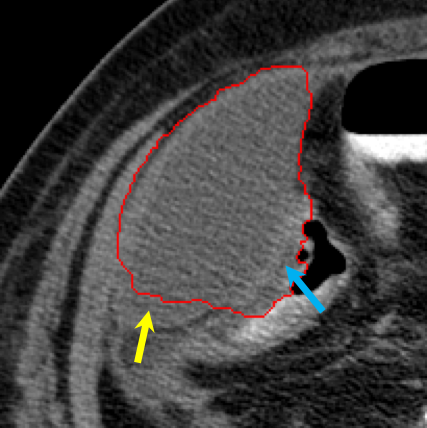}}  \hfill
 \fbox{\includegraphics[width=0.11\linewidth, height=0.11\linewidth]{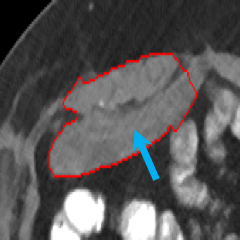}} 
\end{center}
\vspace{-3mm}
\caption{Visual Examples of accepted and rejected lesion segmentation examples from the subjective user study. The \textbf{first row} presents good cases that are accepted the observer. In the \textbf{second row}, failed cases (among rejected segmentations) are shown. The red curves delineate segmented lesion boundaries. The under-segmented lesion regions are illustrated by yellow arrows and the over-segmented healthy tissue areas are indicated by blue arrows. Better viewed in color version.}
\label{fig:visual-example}
\end{figure*}

{\bf Ablation Study of Stacked-GAN:} When we train the HNN using two input channels of original CT and super resolution (SR) images, there are significant improvements in the precision, AVD, VS, and mDICE metrics (see Table.\ref{tab:Ablation-convnet}). In particular, the AVD scores drop by $15\%$ from $0.230$ to $0.189$, likely implying better lesion boundary delineation. Category-wise 2D lesion segmentation results are reported in Table \ref{tab:lesion-2d}. The SGAN assisted HNN (HNN-SR) improves over standard HNN, and, with GrabCut post-processing, the best result of $93\%$ mDICE score is achieved.

\subsection{Subjective Clinical Acceptance Evaluation}
The main motivation and goal of this work is exploring the feasibility of replacing and converting the current 2D RECIST diameter based lesion measurements into 3D volumetric scores. Consequently, it is critical to evaluate clinician's subjective feedback on the acceptance rate of our WSSS produced lesion segmentations. We measure the acceptance rate of a US board-certificated radiologist with over 25 years of clinical practice experience under two different scenarios.

1) The $1,000$ testing RECIST-measured lesion CT images (overlaid with manual or automatic segmentation masks) are randomly displayed to the experienced radiologist to judge whether to accept the segmentation or not. The acceptance rate for WSSS segmentations is $92.4\%$, which is close to the $98.7\%$ acceptance rate of the manually segmented masks (which were performed by 2 trainees under the supervision of a radiologist). Examples of accepted and rejected automatic segmentations are displayed in Fig.~\ref{fig:visual-example}.

2) We show both manual and WSSS computed lesion segmentation results simultaneously in shuffled orders to the radiologist and let him pick the more preferable result (\ie{}, the human judge does not know which segmentation mask is the manual \textit{ground truth}). Consequently, the radiologist picks $72$ \textit{WSSS-over-manual} segmentation cases, designates another $645$ instances as \textit{inseparable} in segmentation quality, and marks the remaining $269$ as \textit{manual-over-WSSS}. 

\begin{figure}[t!]
	\begin{center}
		\includegraphics[width=.80\linewidth]{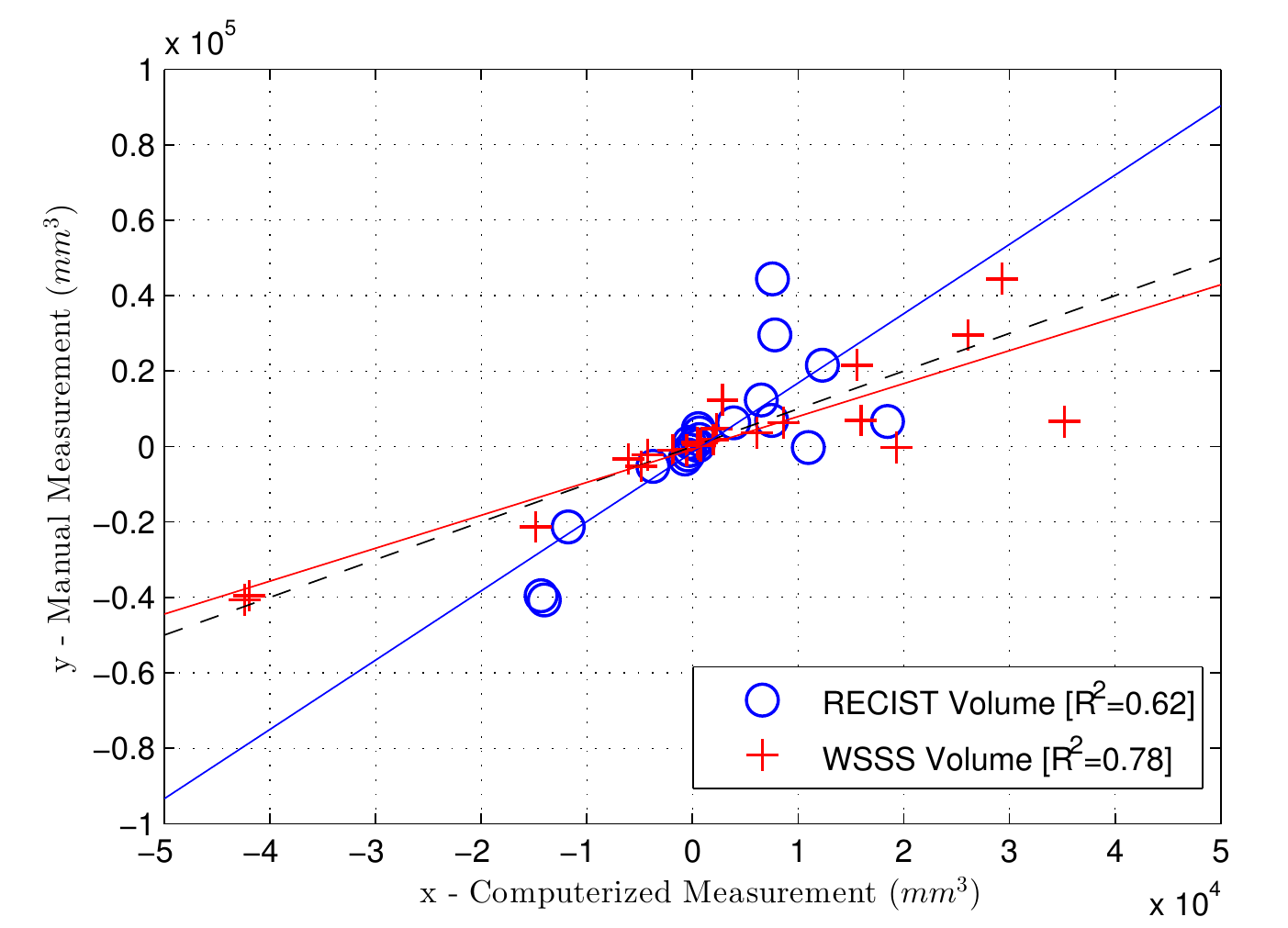}
	\end{center}
	\vspace{-4mm}
	\caption{{\bf Volume Measurements:} Volume changes measured using manual segmentations are graphed vs. WSSS and RECIST-based segmentation. Lines of best fit are also rendered.}
	\label{fig:volume-change}
\end{figure}

Finally, we also measure how well the WSSS method can track lesion changes over time. Toward this end, $22$ pairs of lesion volumes with follow-ups are selected from our test dataset. They are used to evaluate the volume changes at two time points via the means of manual segmentation, RECIST measurements, or the proposed WSSS segmentation. For RECIST, the lesion volume is estimated by ellipsoid volume as $V = \pi (length \times width^2) / 6$ where $length$ and $width$ represent the long and short axes, respectively. In Fig.~\ref{fig:volume-change}, the volume change plots against the follow-up cases are illustrated. Manual and computerized segmentation results correlate well (with $R^2=0.78$), whereas the ellipsoid RECIST estimations tend to report smaller-than-actual volume changes (with $R^2=0.62$).

\section{Discussions \& Conclusion} \label{sec:conclusion}

We present a simple yet surprisingly effective weakly supervised deep segmentation approach on turning massive amounts of RECIST-based lesion diameter measurements (retrospectively stored in hospitals' digital repositories) to full 3D lesion volume segmentation and measurements. The radiologist's \textit{zooming} factor when depicting the lesion diameters on CT images is emulated via our proposed stacked GAN models for image denoising and super-resolution. Importantly, our approach does not require pre-existing RECIST measurements on processing new cases. % of lesion segmentation.

Our method is fully automatic and learned from a large quantity of \textit{partially-labeled} clinical annotations. The lesion segmentation results are validated through both quantitative evaluation (e.g., $93\%$ mean DICE on RECIST-slices and $76\%$ for 3D lesion volume segmentation) and subjective user study. We demonstrate that our self-paced learning improves performance over state-of-the-art CNNs. Moreover, we demonstrate how leveraging the weakly supervised, but large-scale data, allows us to outperform fully-supervised approaches that can only be trained on subsets where full masks are available. Our 3D lesion segmentation also produces more accurate estimations of lesion volume changes than the RECIST criteria. Our work is potentially of high importance for automated and large-scale tumor volume measurement/management in the domain of precision quantitative radiology imaging.

{\small
	\bibliographystyle{ieee}
	\bibliography{BiBTex/egbib_Le}
}

\newpage
\section*{Supplementary Material}
\vspace{1cm}
%-------------------------------------------------------------------------
\begin{figure*}[t!]
	\begin{center}
		\setlength{\fboxsep}{0pt}
		\fbox{\includegraphics[height=0.09\linewidth]{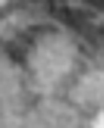}} \hfill
		\fbox{\includegraphics[height=0.09\linewidth]{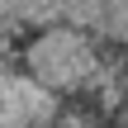}} \hfill
		\fbox{\includegraphics[height=0.09\linewidth]{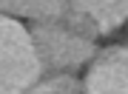}} \hfill
		\fbox{\includegraphics[height=0.09\linewidth]{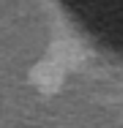}} \hfill
		\fbox{\includegraphics[height=0.09\linewidth]{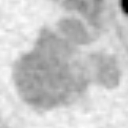}} \hfill
		\fbox{\includegraphics[height=0.09\linewidth]{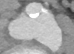}} \hfill
		\fbox{\includegraphics[height=0.09\linewidth]{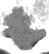}} \hfill
		\fbox{\includegraphics[height=0.09\linewidth]{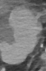}} \hfill
		\fbox{\includegraphics[height=0.09\linewidth]{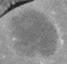}} \hfill
		\fbox{\includegraphics[height=0.09\linewidth]{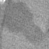}} \\ [1mm]
		\fbox{\includegraphics[height=0.09\linewidth]{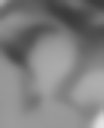}} \hfill
		\fbox{\includegraphics[height=0.09\linewidth]{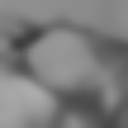}} \hfill
		\fbox{\includegraphics[height=0.09\linewidth]{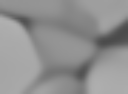}} \hfill
		\fbox{\includegraphics[height=0.09\linewidth]{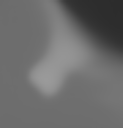}} \hfill
		\fbox{\includegraphics[height=0.09\linewidth]{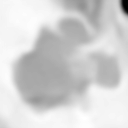}} \hfill
		\fbox{\includegraphics[height=0.09\linewidth]{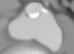}} \hfill
		\fbox{\includegraphics[height=0.09\linewidth]{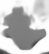}} \hfill
		\fbox{\includegraphics[height=0.09\linewidth]{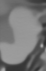}} \hfill
		\fbox{\includegraphics[height=0.09\linewidth]{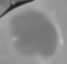}} \hfill
		\fbox{\includegraphics[height=0.09\linewidth]{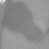}} \\ [1mm]
		\fbox{\includegraphics[height=0.09\linewidth]{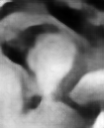}} \hfill
		\fbox{\includegraphics[height=0.09\linewidth]{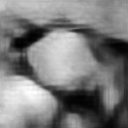}} \hfill
		\fbox{\includegraphics[height=0.09\linewidth]{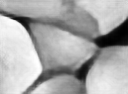}} \hfill
		\fbox{\includegraphics[height=0.09\linewidth]{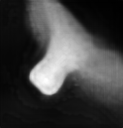}} \hfill
		\fbox{\includegraphics[height=0.09\linewidth]{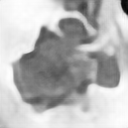}} \hfill
		\fbox{\includegraphics[height=0.09\linewidth]{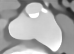}} \hfill
		\fbox{\includegraphics[height=0.09\linewidth]{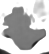}} \hfill
		\fbox{\includegraphics[height=0.09\linewidth]{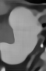}} \hfill
		\fbox{\includegraphics[height=0.09\linewidth]{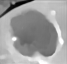}} \hfill
		\fbox{\includegraphics[height=0.09\linewidth]{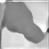}} \hfill
	\end{center}
	\vspace{-4mm}
	\caption{\textbf{Examples of our Stacked GAN based super-resolution approach on lesion images.} The first row illustrates the zoomed images using the bi-cubic interpolation algorithm on original CT lesion images, and the second and third rows demonstrate the corresponding intermediate image de-noising outputs (after the first GAN) and final super resolution results (after the second GAN), respectively.}	\label{fig:sp-examples}
\end{figure*}
\noindent \textbf{Stacked GAN:} In~\cite{ledig2016photo}, generative adversarial networks (GAN) have been successfully used for natural-image super resolution, which produces high-quality images with more low-level visual details and edges from their low-resolution counterparts. For lesion segmentation, if we can improve the visual clarity and contrast of image edges of lesions, the segmentation performance and accuracy may be subsequently improved. As such, our work is partially inspired by \cite{ledig2016photo}. As shown in the first row of Fig.~\ref{fig:sp-examples}, CT images are often noisy and suffer from low contrast due to radiation dosage limits. Directly applying GAN-based super resolution methods on such images generates undesirable visual artifacts and edges that are harmful for lesion segmentation accuracy. To address this problem, the CT imaging noise needs to be reduced before super resolution or spatial zooming. It is challenging to train a single GAN model which directly outputs high resolution images with high visual quality (\eg{}, clear object-level boundaries from noisy images) from the original lesion CT images. Therefore our proposed stacked GAN operates in two stages, breaking the CT-image super resolution process into two sub-tasks: denoising followed by spatial zooming with enhancement.

Given a CT lesion image {\small $X_0$} (as shown in the first row of Fig.~\ref{fig:sp-examples}), we first generate a denoised version of the input image {\small $X_1$} by employing our first GAN model (consisting of a generator {\small $G_1$} and a discriminator {\small $D_1$}) that focuses on removing the random image noise. {\small $X_1$} has the same size as {\small $X_0$}. Although the noise has been reduced in the generated image {\small $X_1$} (as demonstrated in the second row of Fig.~\ref{fig:sp-examples}), lesions have blurry edges and the imaging contrast between lesion and background regions are generally low. However, clear edges and high contrast are desirable and important for achieving high precision lesion segmentation results. As well, a considerable amount of lesions are quite small in size ($<10$mm or less than 10 pixels according to their long axis diameters) and human observers typically apply zooming (via commercial clinical PACS workstations). If we intend to develop an effective CNN model to learn discriminative imaging features for lesion segmentation, image resolutions should be sufficiently higher than the physical CT imaging resolution (approximately in the range of 1mm per pixel). To solve this issue, our second GAN model, which also contains a generator {\small $G_2$} and a discriminator {\small $D_2$}, is built upon the output {\small $X_1$} from the first GAN to produce a high resolution version {\small $X_2$} (as illustrated in the third row of Fig.~\ref{fig:sp-examples}). This high-resolution image provides both clear lesion boundaries and high contrast. Since the three resulting images, \ie{} the original, denoised, and high-resolution variants, as a triplet may have complementary information, we compose them together into a three-channel image that is fed into the next lesion segmentation stage.

We adapt similar architectures as~\cite{ledig2016photo} for the generators and discriminators. In~\cite{ledig2016photo}, the generator has 16 identical residual blocks and 2 sub-pixel convolutional layers~\cite{shi2016real}, which are used to increase the resolution. Each block contains two convolutional layers with 64 $3 \times 3$ kernels followed by batch-normalization~\cite{ioffe2015batch} and ParametricReLU~\cite{he2015delving} layers. For a trained model, the method~\cite{ledig2016photo} can only enlarge the input image by fixed amounts. In the DeepLesion dataset, the lesion sizes vary considerably. Different lesions have to be enlarged with correspondingly different zooming factors. Therefore the sub-pixel layers are removed in the high-resolution generator {\small $G_2$} of our stacked GAN model. Because it is an easier subtask, a simpler architecture that contains just 9 identical residual blocks is designed for the denoising generator {\small $G_1$} of denoising. {\small $G_1$} and {\small $G_2$} are fully convolutional neural networks and can take input images with arbitrary sizes. For the discriminator design, {\small $D_1$} and {\small $D_2$}, we use the same architecture as~\cite{ledig2016photo}, which consists of 8 convolutional layers with $3\times3$ kernels, LeakyReLU activations {\small $(\alpha=0.2)$} and two densely connected layers followed by a final sigmoid layer. The stride settings and numbers of kernels for 8 convolutional layers are {\small $(1,2,1,2,1,2,1,2)$} and {\small $(64,64,128,128,256,256,512,512)$}, respectively. \\

%-------------------------------------------------------------------------
\begin{figure}[h!]
	\centering
	\includegraphics[width=.70\linewidth,trim={3cm 1.5cm 5cm 1.5cm},clip]{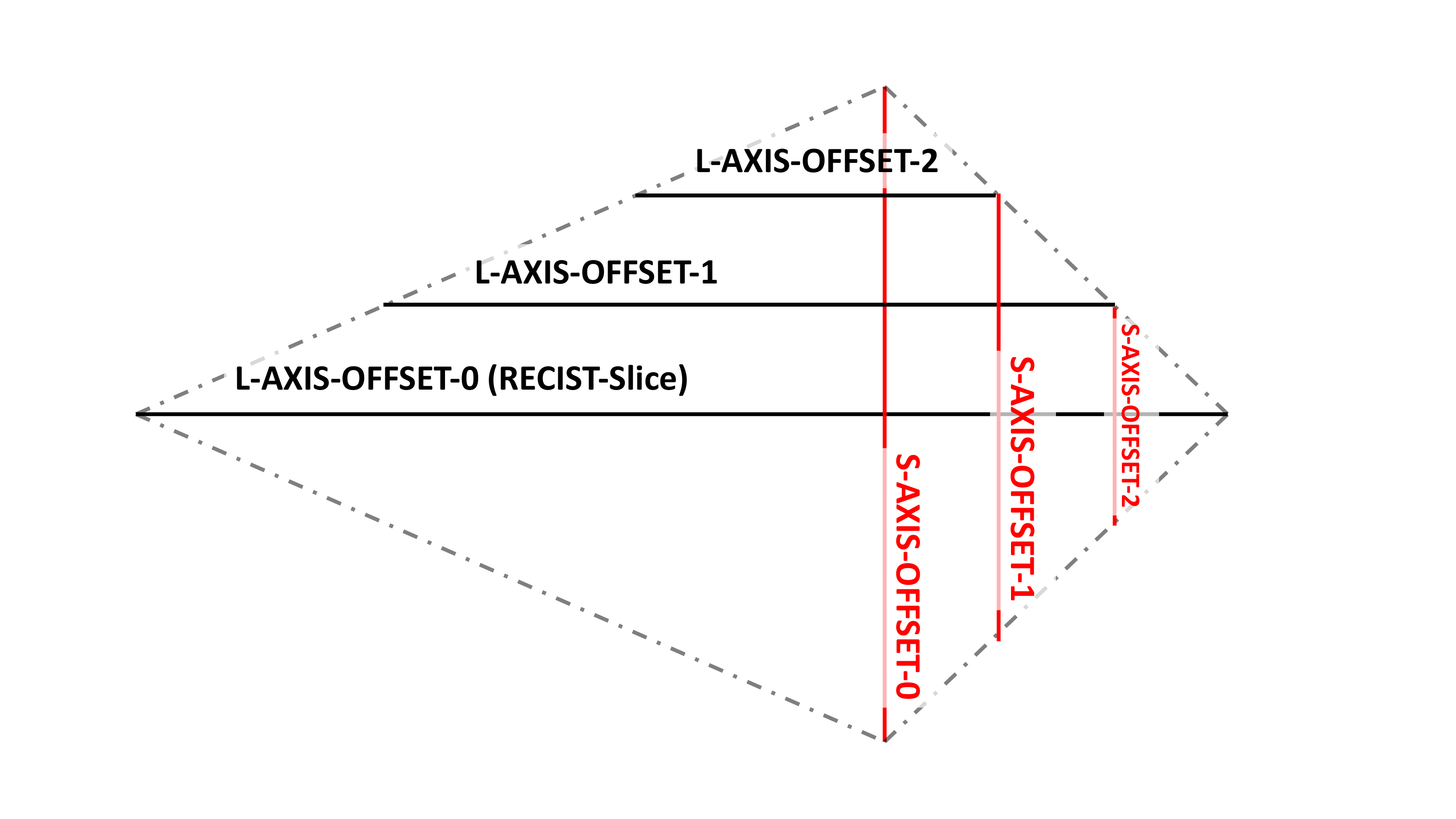}
	\caption{\textbf{3D RECIST Estimation:} L-AXIS-OFFSET-0/1/2 present the lengths of (estimated) long axes on the RECIST/offset-1/offset-2 slices, and S-AXIS-OFFSET-0/1/2 presents the length measurements of their corresponding (estimated) short axes.}
	\label{fig:recist-3d-estimate}
\end{figure}
\noindent \textbf{3D RECIST Estimation:} As described in the main text, the estimated RECIST of off-RECIST-slices is projected from the actual RECIST diameters via Pythagorean theorem using physical Euclidean distance. In Fig.~\ref{fig:recist-3d-estimate}, we visualize the process of RECIST projection, where the length of the long/short axis on the off-RECIST-slice is calculated based on the long/short axis ratio of the actual RECIST diameters. Meanwhile, the intersection of each pair of long and short axes is fixed to the same position as the intersection of the actual RECIST.\\

%-------------------------------------------------------------------------
\begin{table}[t!]
	\begin{center}
		\caption{\textbf{Methods to Generate Training Label from RECIST:} pixel-wise precision, recall, and mean DICE (mDICE) are reported with standard deviation ($\pm$stdv.). Six different setup configurations are compared, 1) RECIST: dilated RECIST, 2) DCRF: dense CRF, 3) GrabCut: uses only RECIST bbox, 4) GrabCut$^i$: uses b-box and interior foreground, 5) GrabCut-R: uses b-box and dilated RECIST, 6) GrabCut-R-SR: implements GrabCut-R on the super-resolution images. See Sec.~\textcolor{red}{3.1} and experiment settings for details.}
		\label{tab:mask-init}
		{\small
			\begin{tabular}{l c c c}
				\hline
				Method & Recall & Precision & mDICE \\
				\hline \hline
				\multicolumn{4}{c}{Deep Lesion RECIST-Slice (Testing Images)} \\
				\hline
				RECIST     & 0.39$\pm$0.13  & \textbf{0.92$\pm$0.14}  & 0.53$\pm$0.14 \\
				DCRF       & 0.72$\pm$0.26  & 0.90$\pm$0.15  & 0.77$\pm$0.20 \\
				GrabCut    & 0.62$\pm$0.46  & 0.68$\pm$0.44  & 0.62$\pm$0.46 \\
				GrabCut$^i$& 0.94$\pm$0.11  & 0.81$\pm$0.16  & 0.86$\pm$0.11 \\
				GrabCut-R  & \textbf{0.94$\pm$0.10}  & 0.89$\pm$0.10  & \textbf{0.91$\pm$0.08} \\
				GrabCut-R-SR & 0.94$\pm$0.11  & 0.90$\pm$0.10  & 0.91$\pm$0.09 \\
				\hline
			\end{tabular}
		}
	\end{center}
\end{table}
\noindent \textbf{Label Map Initializations:} We also consider generating the initial training label maps from the images after super-resolution. As shown in Table~\ref{tab:mask-init}, the row of GrabCut-R-SR presents recall,  precision, and mDICE of the training labels produced from applying GrabCut-R on the super-resolution images. However, we observe that the differences between results produced by GrabCut-R or GrabCut-R-SR are not significant. This may be because GrabCut~\cite{rother_2004_grabcut} cannot fully make use of the augmented image information compared to deep CNN models. Therefore we only use the original CT images to generate training label maps in the main submission. \\

%-------------------------------------------------------------------------
\begin{figure}[t!]
	\centering
	\includegraphics[width=\linewidth]{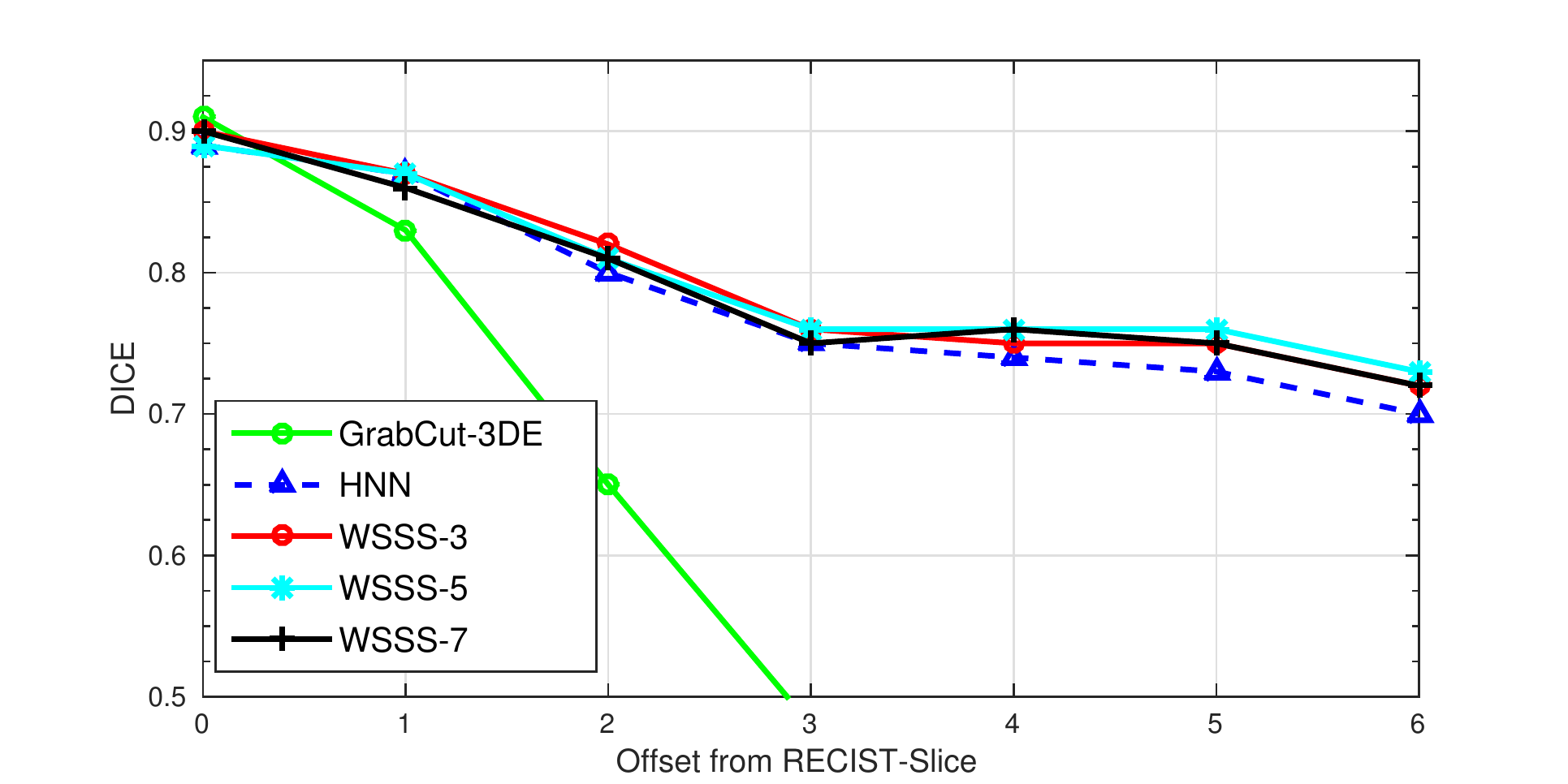}
	\caption{\textbf{Volume Segmentation with Offsets:} The x-axis represents the voxel distance of axial slices to the RECIST-Slice. GrabCut-3DE is GrabCut segmentation performed with 3D RECIST estimation. HNN is trained on the RECIST-slice. WSSS-3/WSSS-5/WSSS-7 is the HNN self-paced with $3$, $5$, and $7$ axial slices, respectively.}
	\label{fig:offsets_3-5-7}
\end{figure}
\begin{figure}[t!]
	\centering
	\mbox{%
		\begin{minipage}[b]{0.49\linewidth}
			\centering
			\scriptsize \hspace{2mm} RECIST-Slice PR-Curve
			\includegraphics[width=\linewidth,height=0.90\linewidth]{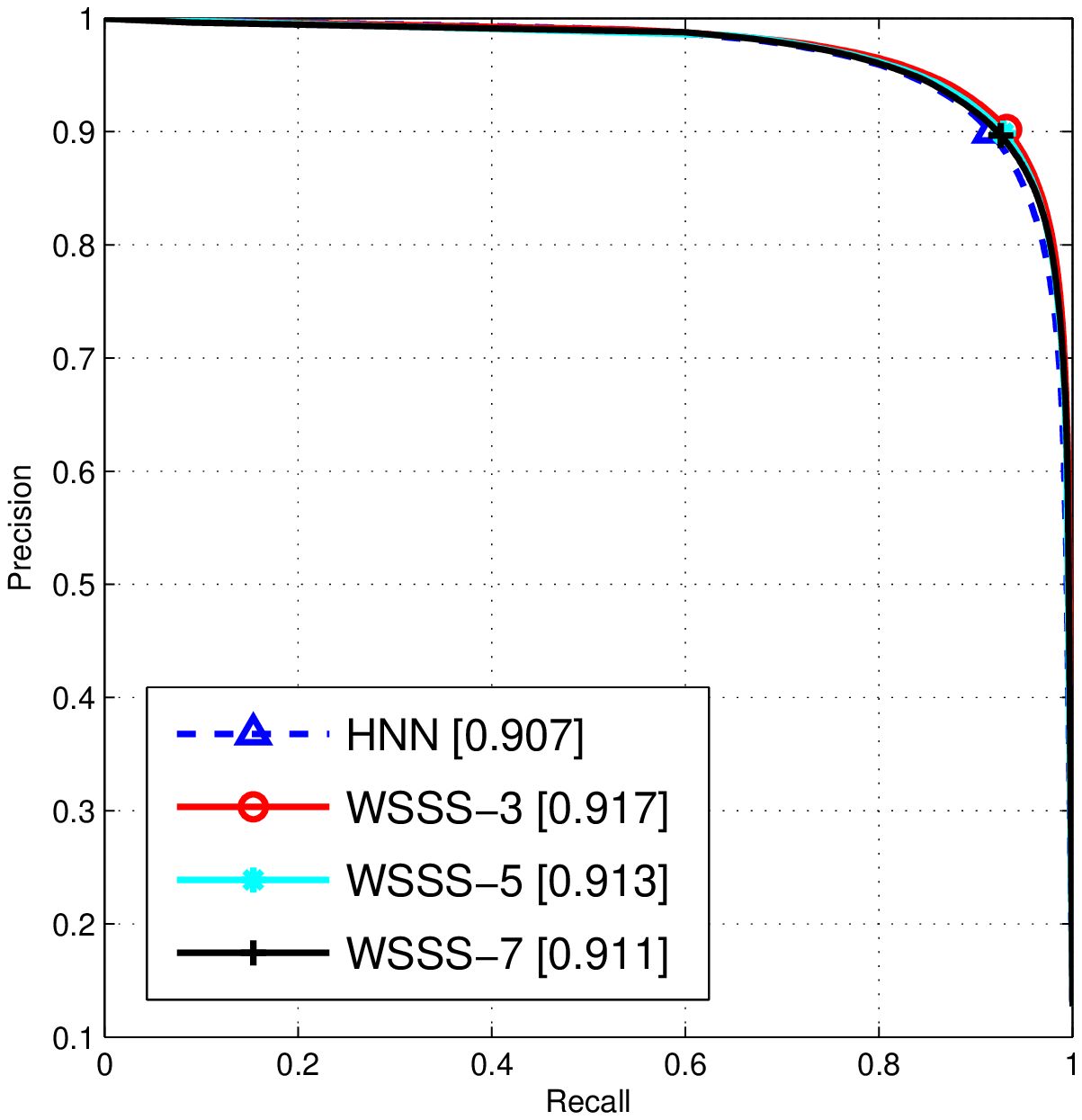}
		\end{minipage}
		\hspace{2mm}
		\begin{minipage}[b]{0.49\linewidth}
			\centering
			\scriptsize \hspace{2mm} Lesion Volume PR-Curve
			\includegraphics[width=\linewidth,height=0.90\linewidth]{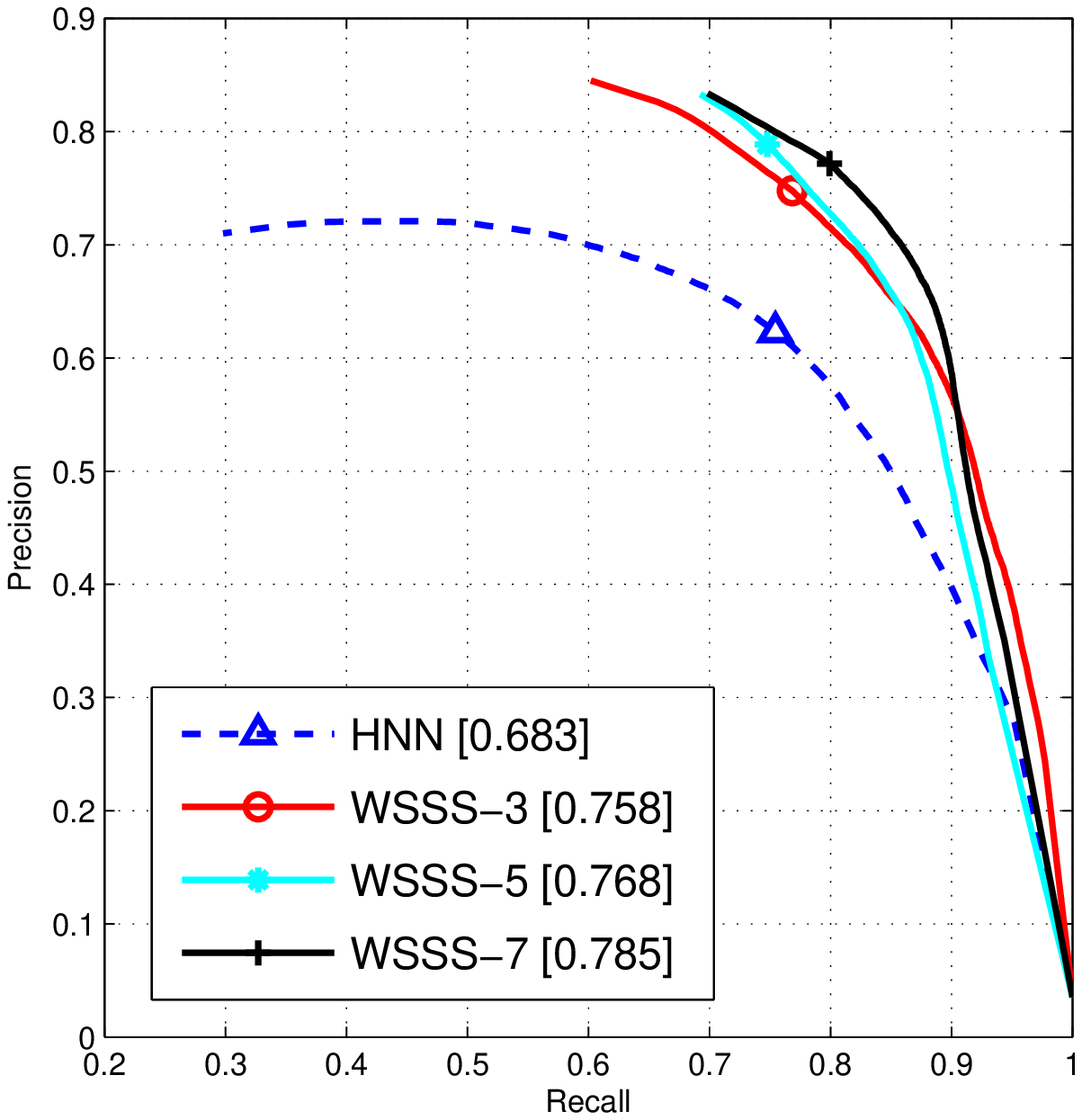}
		\end{minipage} \hspace{3mm}}
	\caption{\textbf{Dataset Precision-Recall:} The precision and recall curves of CNN output on 1,000 testing RECIST slices and 200 lesion volumes. HNN, WSSS-3, WSSS-5, and WSSS-7 share the same definition as in Fig.~\ref{fig:offsets_3-5-7}, and its F-measure is presented in brackets in the legend, respectively.}
	\label{fig:pr_curves_3-5-7}
\end{figure}
\noindent \textbf{3D Segmentation with WSSS:} More quantitative evaluations of the proposed weakly supervised self-paced segmentation (WSSS) are presented in Fig.~\ref{fig:offsets_3-5-7} and Fig.~\ref{fig:pr_curves_3-5-7}. These results are based on the HNN based lesion segmentation models that are trained on the RECIST-slice, and under WSSS-3/5/7 settings, which means that the HNN model is trained in a self-paced manner with 3, 5, and 7 axial slices, respectively. Limited improvement is observed from WSSS-7 against WSSS-5 in Fig.~\ref{fig:offsets_3-5-7}, which indicates the convergence of model training. Therefore, we report the outputs of WSSS-5 as the best 3D segmentation results in the main submission.\\

%-------------------------------------------------------------------------
\begin{figure}[t!]
	\centering
	\includegraphics[width=.90\linewidth]{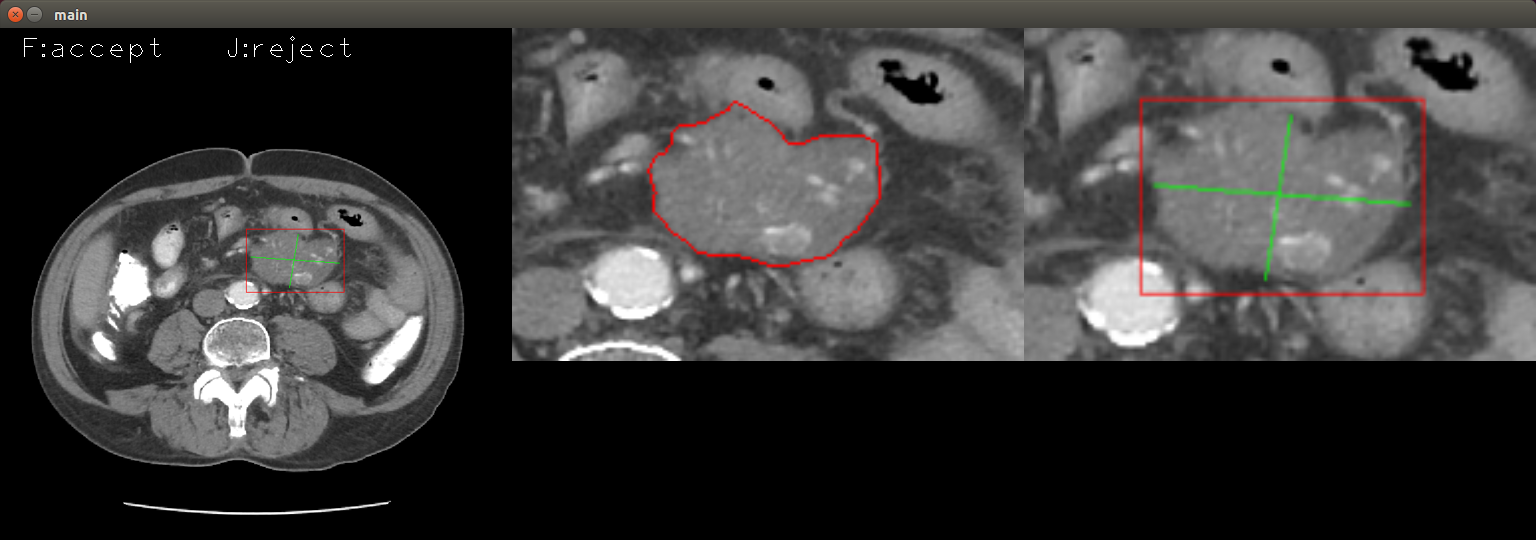}
	\caption{\textbf{GUI for Subjective Acceptance Rate Study of human readers:} The GUI displays the RECIST-slice, zoomed segmentation and RECIST marks, from left to right. The input interface is designed: Key 'f' - To accept, Key 'j' - To reject, Key ',' - To go back, Key 'Esc' - To quit, and Key 'h' - for help.}
	\label{fig:subjective_study_exp1}
\end{figure}
\begin{figure}[t!]
	\centering
	\includegraphics[width=.90\linewidth]{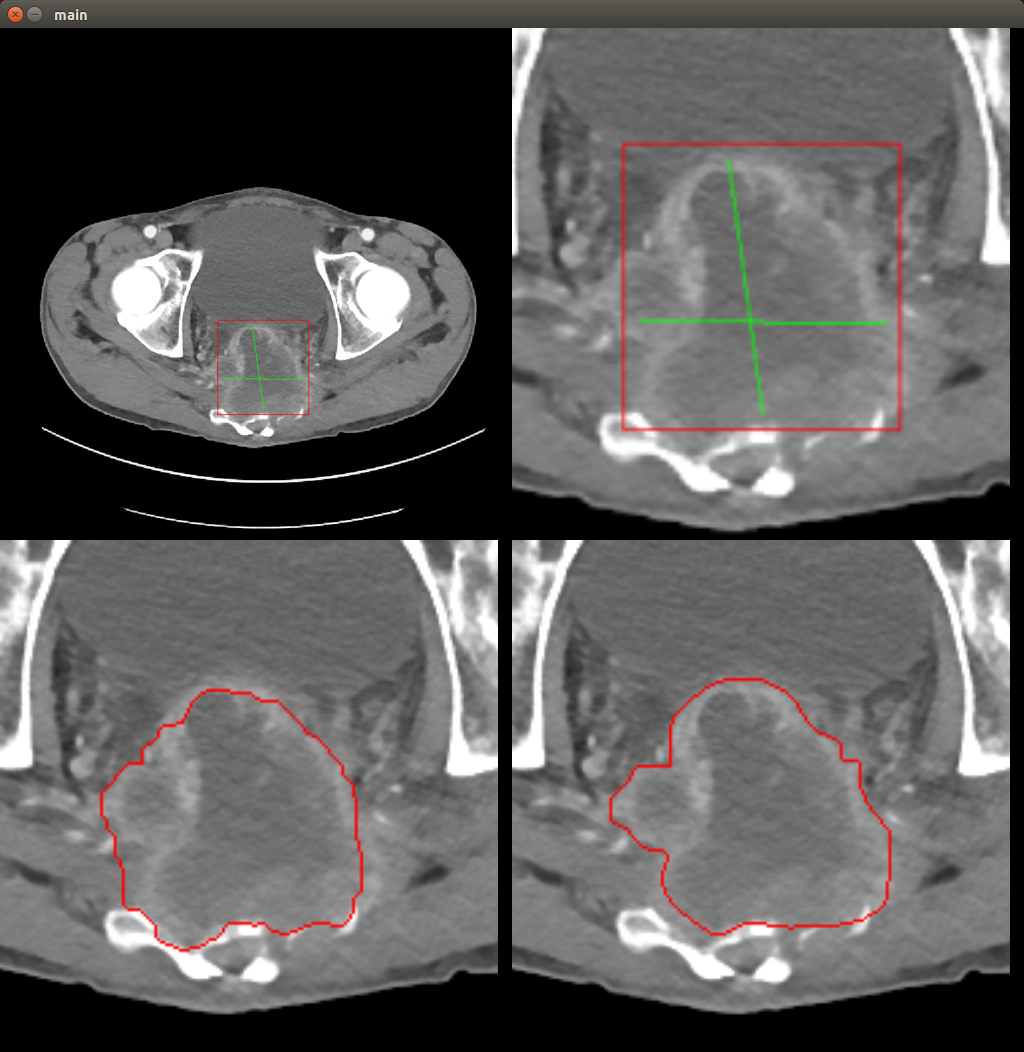}
	\caption{\textbf{GUI for Visual Comparison Study:} The GUI displays the RECIST-slice and zoomed RECIST marks in left and right, respectively, on the first row. On the bottom row, the GUI displays automatic segmentation and manual annotation in a randomly shuffled order. The input interface is designed: Key 'f' - To select left, Key 'j' - To select right, Key 'b' - To select both, Key 'n' - To deny both, Key ',' - To go back, Key 'Esc' - To quit, and Key 'h' - for help.}
	\label{fig:subjective_study_exp2}
\end{figure}
\noindent \textbf{Subjective Study:} GUI snapshots of our subjective study are displayed in Fig.~\ref{fig:subjective_study_exp1} and Fig.~\ref{fig:subjective_study_exp2}. Of note, we allow the radiologist to reject both manual and automatic segmentation in \textit{Visual Comparison Study}, and this explains why the sum of \textit{WSSS-over-manual} (72), \textit{manual-over-WSSS} (269), and \textit{inseparable} (645) is 986, which is less than 1,000. In our study, 14 pairs of both WSSS and manual segmentation results were rejected at the same time. \\

%-------------------------------------------------------------------------
\noindent \textbf{RECIST-Slice Segmentation with Super-resolution CT Images:} Finally in Fig.~\ref{fig:visual-example}, we compare the results of RECIST-slice segmentation with and without image super resolution augmentation via our proposed stacked GAN models. We find, in the lesion categories of abdomen and soft-tissue, the super-resolution images preserve sharper lesion boundaries than the original CT images and their corresponding segmentation results are indeed superior in delineating better lesion boundaries with higher accuracy than raw CT images.
%-------------------------------------------------------------------------
\begin{figure*}[t!]
	\begin{center}
		\setlength{\fboxsep}{0pt}
		\fbox{\includegraphics[height=0.09\linewidth,trim={1mm 4mm 1mm 4mm},clip]{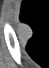}} \hfill
		\fbox{\includegraphics[height=0.09\linewidth,trim={1mm 4mm 1mm 4mm},clip]{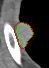}} \hfill
		\fbox{\includegraphics[height=0.09\linewidth,trim={1mm 2mm 1mm 2mm},clip]{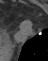}} \hfill
		\fbox{\includegraphics[height=0.09\linewidth,trim={1mm 2mm 1mm 2mm},clip]{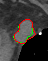}} \hfill
		\fbox{\includegraphics[height=0.09\linewidth,trim={3mm 3mm 3mm 3mm},clip]{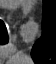}} \hfill
		\fbox{\includegraphics[height=0.09\linewidth,trim={3mm 3mm 3mm 3mm},clip]{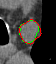}} \hfill
		\fbox{\includegraphics[height=0.09\linewidth,trim={6mm 2mm 6mm 2mm},clip]{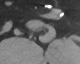}} \hfill
		\fbox{\includegraphics[height=0.09\linewidth,trim={6mm 2mm 6mm 2mm},clip]{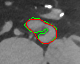}} \hfill
		\fbox{\includegraphics[height=0.09\linewidth,trim={3mm 4mm 3mm 4mm},clip]{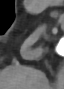}} \hfill
		\fbox{\includegraphics[height=0.09\linewidth,trim={3mm 4mm 3mm 4mm},clip]{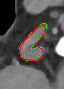}} \hfill
		\fbox{\includegraphics[height=0.09\linewidth,trim={3mm 4mm 3mm 4mm},clip]{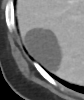}} \hfill
		\fbox{\includegraphics[height=0.09\linewidth,trim={3mm 4mm 3mm 4mm},clip]{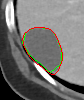}} \\ [1mm]
		\fbox{\includegraphics[height=0.09\linewidth,trim={1mm 4mm 1mm 4mm},clip]{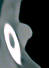}} \hfill
		\fbox{\includegraphics[height=0.09\linewidth,trim={1mm 4mm 1mm 4mm},clip]{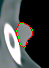}} \hfill
		\fbox{\includegraphics[height=0.09\linewidth,trim={1mm 2mm 1mm 2mm},clip]{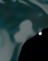}} \hfill
		\fbox{\includegraphics[height=0.09\linewidth,trim={1mm 2mm 1mm 2mm},clip]{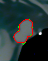}} \hfill
		\fbox{\includegraphics[height=0.09\linewidth,trim={3mm 3mm 3mm 3mm},clip]{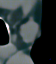}} \hfill
		\fbox{\includegraphics[height=0.09\linewidth,trim={3mm 3mm 3mm 3mm},clip]{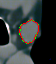}} \hfill		
		\fbox{\includegraphics[height=0.09\linewidth,trim={6mm 2mm 6mm 2mm},clip]{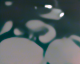}} \hfill
		\fbox{\includegraphics[height=0.09\linewidth,trim={6mm 2mm 6mm 2mm},clip]{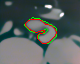}} \hfill
		\fbox{\includegraphics[height=0.09\linewidth,trim={3mm 4mm 3mm 4mm},clip]{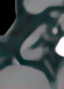}} \hfill
		\fbox{\includegraphics[height=0.09\linewidth,trim={3mm 4mm 3mm 4mm},clip]{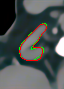}} \hfill
		\fbox{\includegraphics[height=0.09\linewidth,trim={3mm 4mm 3mm 4mm},clip]{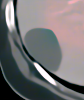}} \hfill
		\fbox{\includegraphics[height=0.09\linewidth,trim={3mm 4mm 3mm 4mm},clip]{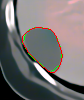}} \\ [4mm]
		\fbox{\includegraphics[height=0.09\linewidth,trim={5mm 2mm 5mm 2mm},clip]{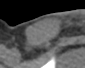}} \hfill
		\fbox{\includegraphics[height=0.09\linewidth,trim={5mm 2mm 5mm 2mm},clip]{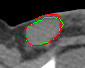}} \hfill
		\fbox{\includegraphics[height=0.09\linewidth]{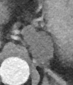}} \hfill
		\fbox{\includegraphics[height=0.09\linewidth]{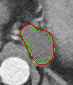}} \hfill
		\fbox{\includegraphics[height=0.09\linewidth,trim={4mm 0 4mm 0},clip]{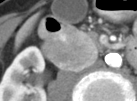}} \hfill
		\fbox{\includegraphics[height=0.09\linewidth,trim={4mm 0 4mm 0},clip]{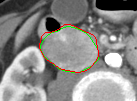}} \hfill
		\fbox{\includegraphics[height=0.09\linewidth,trim={10mm 0 10mm 0},clip]{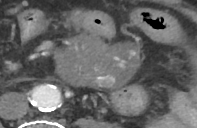}} \hfill
		\fbox{\includegraphics[height=0.09\linewidth,trim={10mm 0 10mm 0},clip]{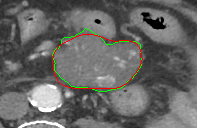}} \hfill
		\fbox{\includegraphics[height=0.09\linewidth]{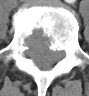}} \hfill
		\fbox{\includegraphics[height=0.09\linewidth]{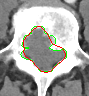}} \\ [1mm]	
		\fbox{\includegraphics[height=0.09\linewidth,trim={5mm 2mm 5mm 2mm},clip]{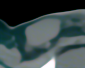}} \hfill
		\fbox{\includegraphics[height=0.09\linewidth,trim={5mm 2mm 5mm 2mm},clip]{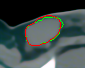}} \hfill
		\fbox{\includegraphics[height=0.09\linewidth]{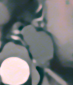}} \hfill
		\fbox{\includegraphics[height=0.09\linewidth]{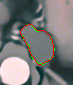}} \hfill
		\fbox{\includegraphics[height=0.09\linewidth,trim={4mm 0 4mm 0},clip]{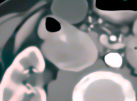}} \hfill
		\fbox{\includegraphics[height=0.09\linewidth,trim={4mm 0 4mm 0},clip]{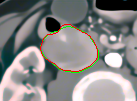}} \hfill			
		\fbox{\includegraphics[height=0.09\linewidth,trim={10mm 0 10mm 0},clip]{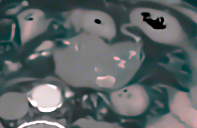}} \hfill
		\fbox{\includegraphics[height=0.09\linewidth,trim={10mm 0 10mm 0},clip]{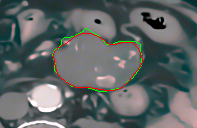}} \hfill
		\fbox{\includegraphics[height=0.09\linewidth]{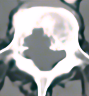}} \hfill
		\fbox{\includegraphics[height=0.09\linewidth]{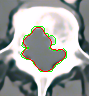}}
	\end{center}
	\vspace{-4mm}
	\caption{\textbf{Lesion segmentation comparison on RECIST-Slices with original images versus super-resolution images.} 11 examples are presented where each example consists of 4 sub-images. In each 4 sub-image group, the top two images are the pair of original CT image and its corresponding HNN segmentation, and the bottom two images are the combined (original+denoised+enhanced) SR-image and the corresponding HNN segmentation. The manual and automatic segmentation are delineated in green and red curves, respectively. Best viewed in color with zooming.}
	\label{fig:visual-example}
\end{figure*}

\end{document}